\documentclass[10pt,twocolumn,letterpaper]{article}

\usepackage{iccv}
\usepackage{times}
\usepackage{epsfig}
\usepackage{graphicx}
\usepackage{amsmath}
\usepackage{amssymb}
\usepackage{booktabs}
\usepackage[skip=1pt]{caption}
\usepackage{subcaption}
\usepackage{algorithm}
\usepackage{listings}
\usepackage{xfrac}
\usepackage[numbers,sort]{natbib}
\usepackage{xspace}
\usepackage{bm}
\usepackage{authblk}
\usepackage{multicol}

\usepackage[pagebackref=true,breaklinks=true,letterpaper=true,colorlinks,bookmarks=false]{hyperref}

\iccvfinalcopy

\newcommand{\methodname}{NNCLR\xspace}

\DeclareMathOperator*{\argmin}{arg\,min}
\newcommand{\norm}[1]{\left\lVert#1\right\rVert}

\ificcvfinal\fi

\begin{document}
\makeatletter
\renewcommand\AB@affilsepx{, \protect\Affilfont}
\makeatother
\title{With a Little Help from My Friends: \\ Nearest-Neighbor Contrastive Learning of Visual Representations}
\author[ 1]{Debidatta Dwibedi}
\author[ 2]{Yusuf Aytar}
\author[ 1]{Jonathan Tompson}
\author[ 1]{Pierre Sermanet}
\author[ 2]{Andrew Zisserman}
\affil[ 1 ]{Google Research}\affil[ 2 ]{DeepMind\protect\\\tt\small \{debidatta, yusufaytar, tompson, sermanet, zisserman\}@google.com}

\maketitle
\ificcvfinal\thispagestyle{empty}\fi

\begin{abstract}
     Self-supervised learning algorithms based on instance discrimination train encoders to be invariant to pre-defined transformations of the \textbf{same} instance.
    While most methods treat different views of the same image as positives for a contrastive loss, we are interested in using positives from \textbf{other} instances in the dataset.
    Our method, Nearest-Neighbor Contrastive Learning of visual Representations (\methodname), samples the nearest neighbors from the dataset in the latent space, and treats them as positives. This provides more semantic variations than pre-defined transformations.

    We find that using the nearest-neighbor as positive in contrastive losses improves performance significantly on ImageNet classification, from 71.7\% to 75.6\%, outperforming previous state-of-the-art methods.
    On semi-supervised learning benchmarks we improve performance significantly when only 1\% ImageNet labels are available, from 53.8\% to 56.5\%. On transfer learning benchmarks our method outperforms state-of-the-art methods (including supervised learning with ImageNet) on 8 out of 12 downstream datasets. Furthermore, we demonstrate empirically that our method is less reliant on complex data augmentations. We see a relative reduction of only 2.1\% ImageNet Top-1 accuracy when we train using only random crops.
\end{abstract}

\vspace*{-2mm}
\section{Introduction}

\begin{figure}[t]
\begin{center}
   \includegraphics[width=0.8\linewidth]{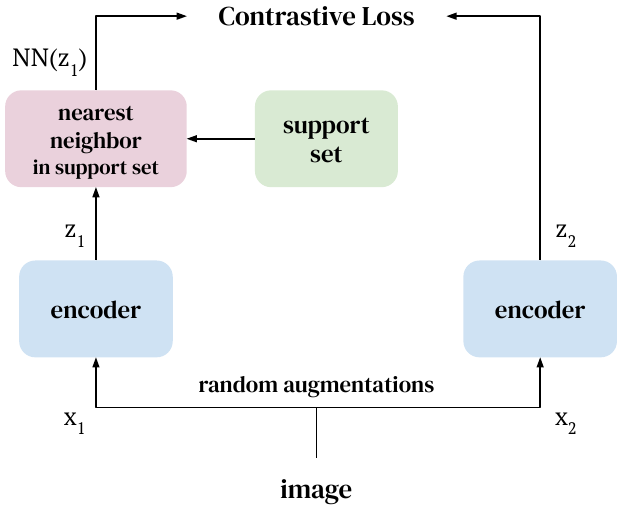}
\end{center}
   \caption{\textbf{\methodname  Training.} We propose a simple self-supervised learning method that uses similar examples from a support set as positives in a contrastive loss.}
\label{fig:teaser}
\end{figure}

How does one make sense of a novel sensory experience? What might be going through someone's head when they are shown a picture of something new, say a dodo? Even without being told explicitly what a dodo is, they will likely form associations between the dodo and other similar semantic classes; for instance a dodo is more similar to a chicken or a duck than an elephant or a tiger. This act of contrasting and comparing new sensory inputs with what one has already experienced happens subconsciously and might play a key role~\cite{GENTNER1999487} in how humans are able to acquire concepts quickly. In this work, we show how an ability to find similarities across items within previously seen examples improves the performance of \textit{self-supervised} representation learning.

A particular kind of self-supervised training -- known as instance discrimination~\cite{wu2018unsupervised,chen2020simple,he2020momentum} -- has become popular recently. Models are encouraged to be invariant to multiple transformations of a \textit{single} sample. This approach has been impressively successful~\cite{grill2020bootstrap,chen2020simple} at bridging the performance gap between self-supervised and supervised models. In the instance discrimination setup, when a model is shown a picture of a dodo, it learns representations by being trained to differentiate between what makes that specific dodo image \textit{different} from everything else in the training set. In this work, we ask the question: if we empower the model to also find other image samples \textit{similar} to the given dodo image, does it lead to better learning?

Current state-of-the-art instance discrimination methods generate positive samples using \emph{data augmentation}, random image transformations (e.g.\ random crops) applied to the same sample to obtain multiple views of the same image. These multiple views are assumed to be positives, and the representation is learned by encouraging the positives to be as close as possible in the embedding space, without collapsing to a trivial solution. However random augmentations, such as random crops or color changes, can not provide positive pairs for different viewpoints, deformations of the same object, or even for other similar instances within a semantic class. The onus of generalization lies heavily on the data augmentation pipeline, which cannot cover all the variances in a given class.

In this work, we are interested in going beyond \emph{single instance positives}, i.e.\ the  instance discrimination task. We expect by doing so we can learn better features that are invariant to different viewpoints, deformations, and even intra-class variations. 
The benefits of going beyond single instance positives have been established in~\cite{Han20,Khosla20}, though these works require class labels or multiple modalities (RGB frames and flow) to obtain the positives which are not applicable to our domain. 
Clustering-based methods~\cite{zhuang2019local,caron2018deep,caron2020unsupervised} also offer an approach to go beyond single instance positives, but assuming the entire cluster (or its prototype) to be positives could hurt performance due to early over-generalization.
Instead we propose using {\em nearest neighbors} in the learned representation space as positives.

We learn our representation by encouraging proximity between different views of the same sample and their nearest neighbors in the latent space.
Through our approach, Nearest-Neighbour Contrastive Learning of visual Representations (\methodname), the model is encouraged to  generalize to new data-points that may not be covered by the data augmentation scheme at hand. In other words, nearest-neighbors of a sample in the embedding space act as small semantic perturbations that are not imaginary, i.e.\ they are representative of actual semantic samples in the dataset. 
We implement our method in a contrastive learning setting similar to~\cite{chen2020simple,chen2020big}. To obtain nearest-neighbors, we utilize a support set that keeps embeddings of a subset of the dataset in memory. This support set also gets constantly replenished during training. 
Note that our support set is different from memory banks~\cite{wu2018unsupervised,tian2019contrastive} and queues~\cite{chen2020improved}, where the stored features are used as negatives. We utilize the support set for nearest neighbor search for retrieving cross-sample positives. Figure~\ref{fig:teaser} gives an overview of the method.

We make the following contributions:
(i) We introduce \methodname to learn self-supervised representations that go beyond single instance positives,
without resorting to clustering;
(ii) We demonstrate that \methodname increases the performance of contrastive learning methods (e.g.\ SimCLR~\cite{chen2020big}) by $\sim 3.8\%$ and achieves state of the art performance on ImageNet classification for linear evaluation and semi-supervised setup with limited labels;
(iii) Our method outperforms state of the art methods on self-supervised, and even supervised features (learned via supervised ImageNet pre-training), on $8$ out of $12$ transfer learning tasks;
Finally, (iv) We show that by using the NN as positive with only random crop augmentations, we achieve $73.3\%$ ImageNet accuracy. This reduces the reliance of self-supervised methods on data augmentation strategies.

\begin{figure*}
     \centering
         \includegraphics[width=\textwidth]{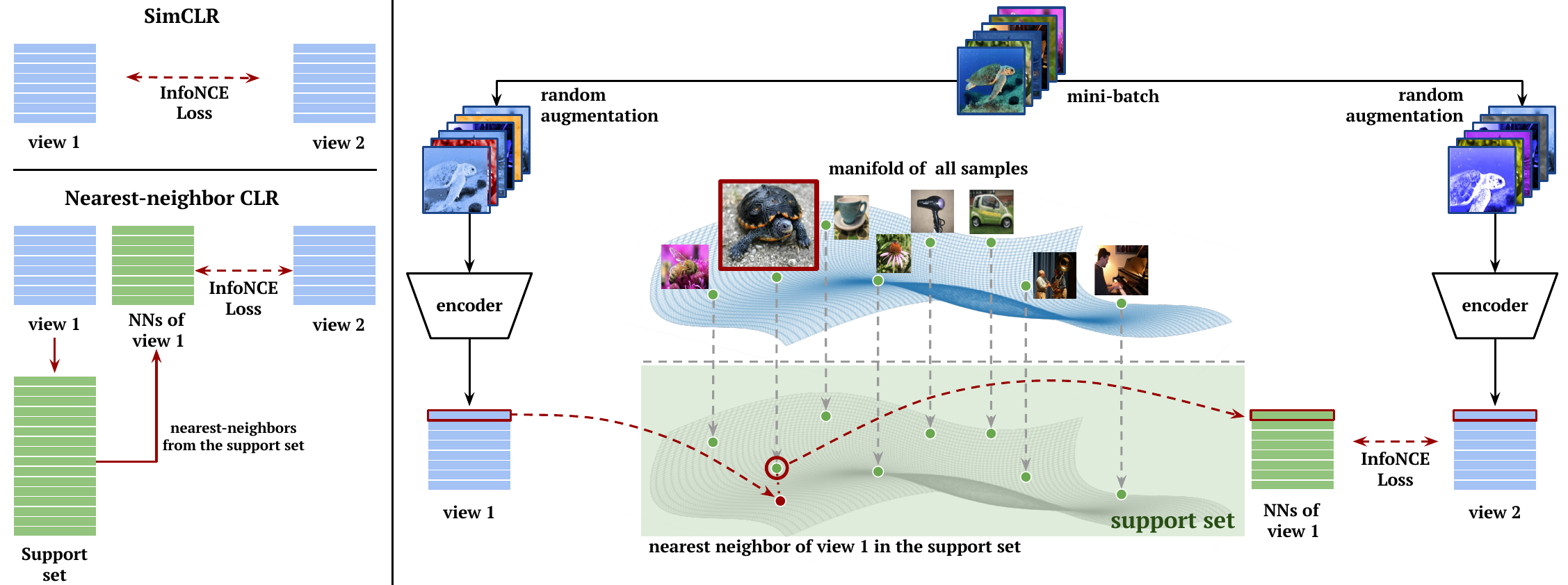}
         \caption{\textbf{Overview of \methodname Training}}
         \label{fig:overview}
\end{figure*}
\section{Related Work}

\noindent \textbf{Self-supervised Learning.} Self-supervised representation learning aims to obtain robust representations of samples from raw data without expensive labels or annotations. Early methods in this field focused on defining \emph{pre-text tasks} -- which typically involves defining a surrogate task on a domain with ample weak supervision labels, like predicting the rotation of images~\cite{gidaris2018unsupervised}, relative positions of patches in an image~\cite{doersch2015unsupervised}, or tracking patches in a video~\cite{wang2015unsupervised, pathak2017learning}. Encoders trained to solve such pre-text tasks are expected to learn general features that might be useful for other downstream tasks requiring expensive annotations (e.g.\ image classification).

One broad category of self-supervised learning techniques are those that use contrastive losses, which have been used in a wide range of computer vision applications\cite{hadsell2006dimensionality, chechik2010large, schroff2015facenet}. These methods learn a latent space that draws positive samples together (e.g.\ adjacent frames in a video sequence), while pushing apart negative samples (e.g.\ frames from another video). In some cases, this is also possible without explicit negatives~\cite{grill2020bootstrap}.
More recently, a variant of contrastive learning called \emph{instance discrimination}~\cite{dosovitskiy2014discriminative, wu2018unsupervised, chen2020simple, chen2020improved} has seen considerable success and have achieved remarkable performance~\cite{chen2020simple, caron2020unsupervised, chen2020big, chen2020improved} on a wide variety of downstream tasks. They have closed the gap with supervised learning to a large extent. Many techniques have proved to be useful in this pursuit: data augmentation, contrastive losses~\cite{chen2020simple, chen2020big,he2020momentum}, momentum encoders~\cite{grill2020bootstrap,chen2020improved,he2020momentum} and memory banks~\cite{chen2020improved,wu2018unsupervised,tian2019contrastive}.
In this work, we extend instance discrimination to include non-trivial positives, not just between augmented samples of the same image, but also from among different images. Methods that use prototypes/clusters~\cite{zhuang2019local,caron2018deep,caron2020unsupervised, bautista2016cliquecnn, asano2019self, caron2019unsupervised, gidaris2020learning, yang2016joint, huang2019unsupervised, xie2016unsupervised, yan2020clusterfit, wu2018improving} also attempt to learn features by associating multiple samples with the same cluster. However, instead of clustering or learning prototypes, we maintain a support set of image embeddings and using nearest neighbors from that set to define positive samples. 

\noindent \textbf{Queues and Memory Banks.} In our work, we use a support set as memory during training. It is implemented as a queue similar to MoCo~\cite{he2020momentum}. MoCo uses elements of the queue as negatives, while this work uses nearest neighbors in the queue to find positives in the context of contrastive losses. \cite{wu2018unsupervised} use a memory bank to keep a running average of embeddings of all the samples in the dataset. Likewise, \cite{zhuang2019local} maintains a clustered set of embeddings and 
uses nearest neighbors to those aggregate embeddings as positives. In our work, the size of the memory is fixed and independent of the training dataset, nor do we perform any aggregation or clustering in our latent embedding space. Instead of a memory bank, SwAV~\cite{caron2020unsupervised} stores prototype centers that it uses for clustering embeddings. SwAV's prototype centers are learned via training with Sinkhorn clustering and persist throughout pre-training. Unlike \cite{wu2018unsupervised, caron2020unsupervised, zhuang2019local}, our support set is continually refreshed with new embeddings and we do not maintain running averages of the embeddings.

\noindent \textbf{Nearest Neighbors in Computer Vision.} Nearest neighbor search has been an important tool across a wide range of computer vision applications~\cite{doersch2012makes,  singh2012unsupervised, hays2008im2gps, hays2007scene, chum2007total, wu2018improving}, from image retrieval to unsupervised feature learning. Nearest-neighbor lookup as an intermediate operation has also been useful for image alignment ~\cite{thewlis2019unsupervised} and video alignment~\cite{dwibedi2019temporal} tasks. \cite{thewlis2019unsupervised} propose a method to learn landmarks on objects in an unsupervised manner by using nearest-neighbors from other images of the same object while \cite{dwibedi2019temporal} show unsupervised learning of action phases by using soft nearest-neighbors across videos of the same action. In our work, we also use cross-sample nearest neighbors but train on datasets with many classes of objects with the objective of learning transferable features. Related to our work, \cite{Han20} uses nearest neighbor retrieval to define self-supervision for video representation learning across different modalities (e.g.\ RGB and optical flow). In contrast, in this work we use nearest neighbor retrieval within a single modality (RGB images), and we maintain an explicit support set of prior embeddings to increase diversity.

\cite{azabou2021mine} concurrently propose leveraging nearest-neighbors in embedding space to improve self-supervised representation learning using BYOL~\cite{grill2020bootstrap}. The authors propose using an additional term for the nearest-neighbor to the BYOL loss. Instead, in our formulation all our loss terms use the nearest-neighbor.
\section{Approach}

We first describe constrastive learning (i.e.\ the InfoNCE loss) in the context of instance discrimination, and discuss SimCLR~\cite{chen2020simple} as one of the leading methods in this domain. Next we introduce our approach, Nearest-Neighbor Contrastive Learning of visual Representations (\methodname), which proposes using nearest-neighbours (NN) as positives to improve contrastive instance discrimination methods. 

\subsection{Contrastive instance discrimination}

\noindent \textbf{InfoNCE}~\cite{sohn2016improved,oord2018representation,wu2018unsupervised} loss (i.e.\ contrastive loss) is quite commonly used in the instance discrimination~\cite{wu2018unsupervised,chen2020simple,he2020momentum} setting. For any given embedded sample $z_i$, we also have another positive embedding $z_i^+$ (often a random augmentation of the sample), and many negative embeddings $z^- \in N_i$. Then the InfoNCE loss is defined as follows:
\begin{equation}\label{eq:infonce}
\mathcal{L}_{i}^\text{InfoNCE} = -  \log{
	{
		\frac{\exp{(z_i\cdot z_i^+/\tau)}}
		{\exp{(z_i \cdot z_i^+ / \tau)}
			+ \sum_{z^- \in \mathcal{N}_i}{\exp{(z_i \cdot z^-/\tau )}}}
	}
} 
\end{equation}
where ($z_i$, $z_i^+$) is the positive pair, ($z_i$, $z^-$) is any negative pair and $\tau$ is the softmax temperature. The underlying idea is learning a representation that pulls positive pairs together in the embedding space, while separating negative pairs.

\noindent \textbf{SimCLR} uses two views of the same image as the positive pair. These two views, which are produced using random data augmentations, 
are fed through an encoder to obtain the positive embedding pair $z_i$ and $z_i^+$. The negative pairs ($z_i$, $z^-$) are formed using all the other embeddings in the given mini-batch. 

Formally, given a mini-batch of images $\{x_1, x_2 .., x_n\}$, two different random augmentations (or views) are generated for each image $x_i$, and fed through the encoder $\phi$ to obtain embeddings $z_{i} = \phi(\text{aug}(x_i))$ and $z_i^+ = \phi(\text{aug}(x_i))$, where $\text{aug}(\cdot)$ is the random augmentation function. The encoder $\phi$ is typically a ResNet-50 with a non-linear projection head. Then the InfoNCE loss used in SimCLR is defined as follows:
\begin{equation}
\label{eqn:simclrloss}
\mathcal{L}_{i}^\text{SimCLR} = -  \log{
	{
		\frac{\exp{(z_i\cdot z_i^+/\tau)}}
		{ \sum\limits_{k=1}^n\exp{(z_i\cdot z_k^+ / \tau)}}
	}
}
\end{equation}
Note that each embedding is $l_2$ normalized before the dot product is computed in the loss.  Then the overall loss for the given mini-batch is $\mathcal{L}^\text{SimCLR}= \frac{1}{n}\sum\limits_{i=1}^n \mathcal{L}^\text{SimCLR}_i$. 

As SimCLR solely relies on transformations introduced by pre-defined data augmentations on the same sample, it cannot link multiple samples potentially belonging to the same semantic class, which in turn might decrease its capacity to be invariant to large intra-class variations. Next we address this point by introducing our method.

\subsection{Nearest-Neighbor CLR (\methodname)}
\label{sec:nnclr}

In order to increase the richness of our latent representation and go beyond single instance positives, we propose using nearest-neighbours to obtain more diverse positive pairs. This requires keeping a \textit{support set} of embeddings which is representative of the full data distribution. 

SimCLR uses two augmentations ($z_i$, $z_i^+$) to form the positive pair. Instead, we propose using $z_i$'s nearest-neighbor in the support set $Q$ to form the positive pair. In Figure~\ref{fig:overview} we visualize this process schematically. Similar to SimCLR we obtain the negative pairs from the mini-batch and utilize a variant of the InfoNCE loss (\ref{eq:infonce}) for contrastive learning. Building upon the SimCLR objective~(\ref{eqn:simclrloss}) we define \methodname loss as below:
\begin{equation}
\label{eqn:nnclrloss}
\mathcal{L}_{i}^\text{NNCLR} = -  \log{
	{
		\frac{\exp{(\text{NN}(z_i, Q)\cdot z_i^+/\tau)}}
		{ \sum\limits_{k=1}^n\exp{(\text{NN}(z_i, Q)\cdot z_k^+ / \tau)}}
	}
}
\end{equation}

where $\text{NN}(z, Q)$ is the nearest neighbor operator as defined below:
\begin{equation}
    \text{NN}(z, Q) = \argmin_{q \in Q} \norm{z - q}_2 \label{eqn:nn}
\end{equation}

As in SimCLR, each embedding is $l_2$ normalized before the dot product is computed in the loss (\ref{eqn:nnclrloss}). Similarly we apply $l_2$ normalization before nearest-neighbor operation in (\ref{eqn:nn}).  We minimize the average loss over all elements in the mini-batch in order to obtain the final loss $\mathcal{L}^\text{NNCLR}= \frac{1}{n}\sum\limits_{i=1}^n \mathcal{L}^\text{NNCLR}_i$. 

\noindent {\bf Implementation details.} We make the loss symmetric~\cite{jia2021scaling,radford2021learning} by adding the following term to Eq.~\ref{eqn:nnclrloss}:  $-\log({{\sfrac{\exp{(\text{NN}(z_i, Q)\cdot z_i^+/\tau)}}{\sum\limits_{k=1}^n\exp{(\text{NN}(z_k, Q)\cdot z_i^+ / \tau)}}}}$
Though, this does not affect performance emperically. Also, inspired from BYOL~\cite{grill2020bootstrap}, we pass $z_i^+$ through a prediction head $g$ to produce embeddings $p_i^+ = g(z_i^+)$. Then we use $p_i^+$ instead of $z_i^+$ in (\ref{eqn:nnclrloss}). Using a prediction MLP adds a small boost to our performance as shown in Section~\ref{sec:ablations}.

\noindent \textbf{Support set.} We implement our support set as a queue (i.e.\ first-in-first-out). The support set is initialized as a random matrix with dimension $[m, d]$, where $m$ is the size of the queue and $d$ is the size of the embeddings. The size of the support set is kept large enough so as to approximate the full dataset distribution in the embedding space. We update it at the end of each training step by taking the $n$ (batch size) embeddings from the current training step and concatenating them at the end of the queue. We discard the oldest $n$ elements from the queue. We only use embeddings from one view to update the support set. Using both views' embeddings to update does not lead to any significant difference in downstream performance. In Section~\ref{sec:ablations} we compare the performance of multiple support set variants.

\section{Experiments}

In this section we compare \methodname features with other state of the art self-supervised image representations. First, we provide details of our architecture and training process. Next, following commonly used evaluation protocol~\cite{grill2020bootstrap,chen2020simple,chen2020big,he2020momentum},
we compare our approach with other self-supervised features on linear evaluation and semi-supervised learning on the ImageNet ILSVRC-2012 dataset. Finally we present results on transferring self-supervised features to other downstream datasets and tasks.

\subsection{Implementation details}

\noindent\textbf{Architecture.} We use ResNet-50~\cite{he2016deep} as our encoder to be consistent with the existing literature ~\cite{grill2020bootstrap,chen2020simple}.
We spatially average the output of ResNet-50 which makes the output of the encoder a 2048-d embedding. The architecture of the projection MLP is $3$ fully connected layers of sizes $[2048, 2048, d]$ where $d$ is the embedding size used to apply the loss. We use $d=256$ in the experiments unless otherwise stated. All fully-connected layers are followed by  batch-normalization~\cite{ioffe2015batch}. All the batch-norm layers except the last layer are followed by ReLU activation. The architecture of the prediction MLP $g$ is $2$ fully-connected layers of size $[4096, d]$.  The hidden layer of the prediction MLP is followed by batch-norm and ReLU. The last layer has no batch-norm or activation.

\noindent\textbf{Training.} Following other self-supervised methods~\cite{grill2020bootstrap,chen2020simple,chen2020big,he2020momentum}, we train our \methodname representation on the ImageNet2012 dataset which contains $1,281,167$ images, without using any annotation or class labels. We train for $1000$ epochs with a warm-up of $10$ epochs with cosine annealing schedule using the LARS optimizer~\cite{you2017large}. Weight-decay of $10^{-6}$ is applied during training. As is common practice~\cite{grill2020bootstrap,chen2020big}, we don't apply weight-decay to the bias terms. We use the data augmentation scheme used in BYOL~\cite{grill2020bootstrap} and we use a temperature $\tau$ of $0.1$ when applying the softmax during computation of the contrastive loss in Equation~\ref{eqn:nnclrloss}. The best results of \methodname are achieved with $98,304$ queue size and base learning rate~\cite{grill2020bootstrap} of $0.3$.

\subsection{ImageNet evaluations}

\noindent\textbf{ImageNet linear evaluation.} Following the standard linear evaluation procedure~\cite{grill2020bootstrap,chen2020simple} we train a linear classifier for $90$ epochs on the frozen $2048$-d embeddings from the ResNet-50 encoder using LARS with cosine annealed learning rate of $1$ with Nesterov momentum of $0.9$ and batch size of $4096$. 

Comparison with state of the art methods is presented in Table~\ref{tab:main_results}. First, \methodname achieves the best performance compared to all the other methods using a ResNet-50 encoder trained with two views. \methodname provides more than $3.6\%$ improvement over well known constrastive learning approaches such as MoCo v2~\cite{chen2020improved} and SimCLR v2~\cite{chen2020big}. Even compared to InfoMin Aug.\ \cite{tian2020makes}, which explicitly studies ``good view'' transformations to apply in contrastive learning, \methodname achieves more than $2\%$ improvement on top-1 classification performance. We outperform BYOL\cite{grill2020bootstrap} (which is the state-of-the-art method among methods that use two views) by more than $1\%$.
 
We also achieve $3.6\%$ improvement compared to the state of the art clustering based method SwAV~\cite{caron2020unsupervised} in the same setting of using two views. 
To compare with SwAV's multi-crop models, we pre-train for $800$ epochs with $8$ views (two $224\times224$ and six $96\times96$ views) using only the larger views to calculate the NNs. In this setting our method outperforms SwAV by $0.3\%$ in Top-1 accuracy. Note that while multi-crop is responsible for $3.5\%$ performance improvement for SwAV, for our method it provides a boost of only $0.2\%$.
However, increasing the number of crops quadratically increases the memory and compute requirements, and is quite costly even when low-resolution crops are used as in~\cite{caron2020unsupervised}.

\noindent\textbf{Semi-supervised learning on ImageNet.} We evaluate the effectiveness of our features in a semi-supervised setting on ImageNet $1\%$ and $10\%$ subsets following the standard evaluation protocol~\cite{grill2020bootstrap, chen2020big}. 
We present these results in Table~\ref{tab:semi_supervised_learning}. The first key result of Table~\ref{tab:semi_supervised_learning} is that our method outperforms all the state of the art methods on semi-supervised learning on ImageNet $1\%$ subset, including SwAV's~\cite{caron2020unsupervised} multi-crop setting.
This is a clear indication of good generalization capacity of \methodname features, particularly in low-shot learning scenarios. Using the ImageNet $10\%$ subset, \methodname outperforms SimCLR~\cite{chen2020simple} and other methods. However, SwAV's~\cite{caron2020unsupervised} multi-crop setting outperforms our method in ImageNet $10\%$ subset.

\begin{table}[]
    \centering
    \begin{tabular}{l|cc}
         Method & Top-1 & Top-5\\
         \midrule

         PIRL~\cite{misra2020self} & 63.6 & - \\
         CPC v2~\cite{henaff2020data} & 63.8 & 85.3 \\
         PCL~\cite{li2020prototypical} & 65.9 & - \\         
         CMC~\cite{tian2019contrastive} & 66.2 & 87.0 \\
          
         MoCo v2~\cite{chen2020improved} & 71.1  & - \\
         SimSiam~\cite{chen2020exploring} & 71.3  & - \\
         SimCLR v2~\cite{chen2020big} & 71.7 & - \\
         SwAV~\cite{caron2020unsupervised}  & 71.8 & - \\
         InfoMin Aug.~\cite{tian2020makes} & 73.0 & 91.1\\
         BYOL~\cite{grill2020bootstrap} & 74.3 & 91.6\\ 

         \methodname (ours) & \textbf{75.4} & \textbf{92.3} \\
         \midrule
         SwAV (multi-crop)~\cite{caron2020unsupervised} & 75.3 & - \\  
         \methodname (ours) (multi-crop) & \textbf{75.6} & \textbf{92.4}\\  
        
    \end{tabular}
    \caption{{\bf ImageNet linear evaluation results.} Comparison with other self-supervised learning methods on ResNet-50 encoder. Methods on the top section use two views only.
    }
    \label{tab:main_results}
\end{table}

\begin{table}[]
    \centering
    \begin{tabular}{l|c c|c c}
          & \multicolumn{2}{c|}{ImageNet \bf{1\%}} &  \multicolumn{2}{c}{ImageNet \bf{10\%}} \\
         Method & Top-1 & Top-5 & Top-1 & Top-5 \\
         \midrule
         Supervised & 25.4 & 48.4 & 56.4 & 80.4\\
         \midrule
         InstDisc~\cite{wu2018unsupervised} & - & 39.2 & - & 77.4\\
         PIRL~\cite{misra2020self} & -  & 57.2 & - & 83.8\\
         PCL~\cite{li2020prototypical} & - & 75.6 & - & 86.2\\ 
         SimCLR~\cite{chen2020simple} & 48.3 & 75.5 & 65.6 & 87.8\\
         BYOL~\cite{grill2020bootstrap} & 53.2  & 78.4 & 68.8 & 89.0\\

         \methodname (ours) & \textbf{56.4} & \textbf{80.7} & \textbf{69.8} & \textbf{89.3} \\
         \midrule SwAV (multi-crop)~\cite{caron2020unsupervised}& 53.9  & 78.5 & 70.2 & 89.9
    \end{tabular}
    \caption{{\bf Semi-supervised learning results on ImageNet.} Top-1 and top-5 performances are reported on fine-tuning a pre-trained ResNet-50 with ImageNet $1\%$ and $10\%$ datasets. }
    \label{tab:semi_supervised_learning}
\end{table}

\begin{table*}[]
\footnotesize
    \centering
    \begin{tabular}{l|cccccccccccc}
         Method & Food101 & CIFAR10 & CIFAR100 &  Birdsnap & SUN397 &  Cars & Aircraft & VOC2007 &  DTD & Pets & Caltech-101 & Flowers\\
         \midrule
         BYOL~\cite{grill2020bootstrap} &  75.3 & 91.3 &  78.4  & 57.2 & 62.2   & \textbf{67.8} &  60.6& 82.5 & 75.5 & 90.4 & 94.2 & \textbf{96.1} \\
         SimCLR~\cite{grill2020bootstrap} & 72.8 & 90.5 & 74.4 & 42.4 & 60.6 & 49.3  & 49.8 & 81.4 & \textbf{75.7} & 84.6 & 89.3 & 92.6 \\
         Sup.-IN~\cite{chen2020simple} & 72.3 & 93.6 & 78.3 &53.7 & 61.9 & 66.7 & 61.0& 82.8 & 74.9 & 91.5 & \textbf{94.5} & 94.7 \\
         \methodname & \textbf{76.7}  & \textbf{93.7} & \textbf{79.0}  & \textbf{61.4} & \textbf{62.5}  & 67.1 & \textbf{64.1} & \textbf{83.0} & 75.5  & \textbf{91.8} & 91.3 & 95.1\\
         
    \end{tabular}
    \caption{\textbf{Transfer learning performance} using ResNet-50 pretrained with ImageNet. For all datasets we report Top-1 classification accuracy except Aircraft, Caltech-101, Pets and Flowers for which we report mean per-class accuracy and VOC2007 for which we report 11-point MAP.}
    \label{tab:transfer_learning}
\end{table*}

\subsection{Transfer learning evaluations}
\label{sec:transfer_learning}
We show representations learned using \methodname are effective for transfer learning on multiple downstream classification tasks on a wide range of datasets. We follow the linear evaluation setup described in~\cite{grill2020bootstrap}. The datasets used in this benchmark are as follows: Food101~\cite{bossard14}, CIFAR10~\cite{Krizhevsky09learningmultiple}, CIFAR100~\cite{Krizhevsky09learningmultiple}, Birdsnap~\cite{berg-birdsnap-cvpr2014}, Sun397~\cite{Xiao:2010}, Cars~\cite{KrauseStarkDengFei-Fei_3DRR2013}, Aircraft~\cite{maji13fine-grained}, VOC2007~\cite{pascal-voc-2007}, DTD~\cite{cimpoi14describing}, Oxford-IIIT-Pets~\cite{parkhi12a}, Caltech-101~\cite{FeiFei2004LearningGV} and Oxford-Flowers~\cite{Nilsback08}. Following the evaluation protocol outlined in~\cite{grill2020bootstrap}, we first train a linear classifier using the training set labels while choosing the best regularization hyper-parameter on the respective validation set. Then we combine the train and validation set to create the final training set which is used to train the linear classifier that is evaluated on the test set. 

We present transfer learning results in Table~\ref{tab:transfer_learning}. \methodname outperforms supervised features (ResNet-50 trained with ImageNet labels) on 11 out of the 12 datasets. Moreover our method improves over BYOL~\cite{grill2020bootstrap} and SimCLR~\cite{chen2020simple} on 8 out of the 12 datasets. These results further validate the generalization performance of \methodname features.

\subsection{Ablations}
\label{sec:ablations}
In this section we present a thorough analysis of \methodname. 
After discussing the default settings, we start by demonstrating the effect of training with nearest-neighbors in a variety of settings. 
Then, we present several design choices such as support set size, varying k in top-k nearest neighbors, type of nearest neighbors, different training epochs, variations of batch size, and embedding size. 
We also briefly discuss memory and computational overhead of our method.

\noindent \textbf{Default settings.} Unless otherwise stated our support set size during ablation experiments is $32,768$ and our batch size is $4096$.  
We train for 1000 epochs with a warm-up of 10 epochs, base learning rate of $0.15$ and cosine annealing schedule using the LARS optimizer~\cite{you2017large}. We also use the prediction head by default. All the ablations are performed using the ImageNet linear evaluation setting.

\noindent \textbf{Nearest-neighbors as positives.} Our core contribution in this paper is using nearest-neighbors (NN) as positives
in the context of contrastive self-supervised learning.
Here we investigate how this particular change, using nearest neighbors as positives, affects performance in various settings with and without momentum encoders. This analysis is presented in Table~\ref{tab:nn_ablation}. First we show using the NNs in contrastive learning (row 2) is $3\%$ better in Top-1 accuracy than using view 1 embeddings (similar to SimCLR) shown in row 1. We also explore using momentum encoder (similar to MoCo~\cite{he2020momentum}) in our contrastive setting. Here using NNs also improves the top-1 performance by $2.4\%$. 

\begin{table}[]
    \centering
    \begin{tabular}{c|c|cc}
         Mom. Enc. & Positive & Top-1  & Top-5 \\
         \midrule
    & View 1 & 71.4 & 90.4\\

           & NN of View 1 & \textbf{74.5} & \textbf{91.9} \\
         \midrule
          \checkmark & View 1 &  72.5 & 91.3 \\
          \checkmark & NN of View 1  & \textbf{74.9} & \textbf{92.1}\\
    \end{tabular}
    \caption{\textbf{Effect of adding nearest-neighbors as positives} in various settings. Results are obtained for ImageNet linear evaluation.}
    \label{tab:nn_ablation}
\end{table}

\begin{table}[]
    \centering
    \begin{tabular}{c|l|l|l}
         Method & SimCLR~\cite{chen2020simple} & BYOL~\cite{grill2020bootstrap} & \methodname\\
         \midrule
         Full aug. & 67.9 & 72.5  & 72.9\\
         Only crop & 40.3 \footnotesize{($\downarrow$ -27.6)}& 59.4 \footnotesize{($\downarrow$ -13.1)} & \textbf{68.2} \footnotesize{($\downarrow$ -4.7)}\\
    \end{tabular}
    \caption{\textbf{Performance with only crop augmentation.} ImageNet top-1 performance for linear evaluation is reported.}
    \label{tab:data_augmentation}
\end{table}

\noindent \textbf{Data Augmentation.} Both SimCLR~\cite{chen2020simple} and BYOL~\cite{grill2020bootstrap} rely heavily on a well designed data augmentation pipeline to get the best performance. However, \methodname is less dependent on complex augmentations as nearest-neighbors already provide richness in sample variations. In this experiment, we remove all color augmentations and Gaussian blur, and train with random crops as the only method of augmentation for 300 epochs following the setup used in~\cite{grill2020bootstrap}. We present the results in Table~\ref{tab:data_augmentation}. We notice \methodname achieves $68.2\%$ top-1 performance on the ImageNet linear evaluation task suffering a performance drop of only $4.7\%$. On the other hand, SimCLR and BYOL suffer larger relative drops in performance, $27.6\%$ and $13.1\%$ respectively. The performance drop reduces further as we train our approach longer. With $1000$ pre-training epochs, \methodname with all augmentations achieves $74.9\%$ while with only random crops \methodname manages to get $73.3\%$, further reducing the gap to just $1.6\%$.  While \methodname also benefits from complex data augmentation operations, the reliance on color jitter and blurring operations is much less. This is encouraging for adopting \methodname for pre-training in domains where data transformations used for ImageNet might not be suitable.

\begin{table}[]
    \centering
    \begin{tabular}{l|cccc}
         Method & 100 & 200 & 400 & 800\\
         \midrule
         SimCLR~\cite{chen2020simple} & 66.5 & 68.3 & 69.8 & 70.4 \\
         MoCov2~\cite{chen2020improved} & 67.4 & 69.9 & 71.0 &  72.2\\
         BYOL~\cite{grill2020bootstrap} & 66.5& 70.6& 73.2 & 74.3 \\
         SWAV~\cite{caron2020unsupervised} & 66.5 & 69.1 & 70.7 & 71.8\\
         SimSiam~\cite{chen2020exploring} & 68.1 & 70.0 & 70.8 & 71.3\\
         \midrule
         \methodname & \textbf{69.4} & \textbf{70.7} & \textbf{74.2} & \textbf{74.9}  \\

    \end{tabular}
    \caption{\textbf{Number of pre-training epochs vs.\ performance.} Results are obtained for ImageNet linear evaluation.}
    \label{tab:pretraining_spochs}
\end{table}
\noindent \textbf{Pre-training epochs.} In Table~\ref{tab:pretraining_spochs}, we show how our method compares to other methods when we have different pre-training epoch budgets. \methodname is better than other self-supervised methods when pre-training budget is kept constant. We find that base learning rate of $0.4$ works best for $100$ epochs, and $0.3$ works for $200$, $400$ and $800$ epochs. 

\noindent \textbf{Support set size.} Increasing the size of the support set increases performance in general. We present results of this experiment in Table~\ref{tab:queuesize}.  
By using a larger support set, we increase the chance of getting a closer nearest-neighbour from the full dataset. As also shown in Table~\ref{tab:topk}, getting the closest (i.e.\ top-1) nearest-neighbour obtains the best performance, even compared against top-2. 

We also find that increasing the support set size beyond $98,304$ doesn't lead to any significant increase in performance possibly due to an increase in the number of stale embeddings in the support set.

\noindent \textbf{Nearest-neighbor selection strategy.} 
Instead of using the nearest-neighbor, we also experiment with taking one of the top-$k$ NNs randomly. These results are presented in Table~\ref{tab:topk}. Here we investigate whether increasing the diversity of nearest-neighbors (i.e.\ increasing $k$) results in improved performance. 
Although our method is somewhat robust to changing the value of $k$, we find that increasing the top-$k$ beyond $k=1$ always results in slight degradation in performance. 
Inspired by recent work~\cite{dwibedi2019temporal, frosst2019analyzing} we also investigate using a soft-nearest neighbor, a convex combination of embeddings in the support-set where each one is weighted by its similarity to the embedding (see~\cite{dwibedi2019temporal} for details). We present results in Table~\ref{tab:hard_vs_soft}. We find that the soft nearest neighbor can be used for training but results in worse performance than using the hard NN. 

\noindent \textbf{Batch size.} Batch size has shown to be an important factor that affects performance, particularly in the contrastive learning setting. We vary the batch size and present the results in Table~\ref{tab:batch_size}. In general, larger batch sizes improve the performance peaking at $4096$.

\noindent \textbf{Embedding size.} Our method is robust to choice of the embedding size as shown in Table~\ref{tab:embedding_size}. We vary the embedding size in powers of 2 from 128 to 2048 and find similar performance over all settings.

\noindent \textbf{Prediction head} As shown in Table~\ref{tab:predhead}, adding a prediction head results in a modest $0.4\%$ boost in the top-1 performance. 

\noindent \textbf{Different implementations of support set.}
We also investigate some variants of how we can implement the support set from which we sample the nearest neighbor. We present results of this experiment in Table~\ref{tab:memory_variations}. In the first row, instead of using a queue, we pass a random set of images from the dataset through the current encoder and use the nearest neighbor from that set of embeddings. This works reasonably well but we are limited by how many examples we can fit in accelerator memory. Since we cannot increase the size of this set beyond $16384$, which results in sub-optimal performance. 
Also using a random set of size $16384$ is about four times slower than using a queue (when training with a batch size of $4096$) as each forward pass requires four times more samples through the encoder. This experiment also shows that \methodname does not need features of past samples (akin to momentum encoders ~\cite{he2020momentum}) to learn representations. We also experiment with updating the elements in the support set randomly as opposed to the default FIFO manner. We find that FIFO results in more than $2\%$ better ImageNet linear evaluation Top-1 accuracy.

\begin{table*}[h]
\begin{subtable}[h]{0.5\textwidth}
    \centering
    \begin{tabular}{c|ccccc}
         Queue Size & 8192 & 16384 & 32768 & 65536 & 98304 \\
         \midrule
         Top-1 & 73.6 &  74.2 & 74.9 &  75.0 & \textbf{75.4}\\
         Top-5 & 91.2 &  91.7 & 92.1 &  92.2 & \textbf{92.3}\\
 
    \end{tabular}
    \caption{Support set size}
    \label{tab:queuesize}
   \end{subtable}
   \hfill
  \begin{subtable}[h]{0.5\textwidth}
    \centering
    \begin{tabular}{c|cccccc}
       \textit{k} in \textit{k}-NN  & 1& 2 & 4 & 8 & 16 & 32\\
         \midrule
         Top-1 & \textbf{74.9} & 74.1 & 73.8 &  73.8 & 73.8 & 73.2\\
         Top-5 & \textbf{92.1} & 91.6 & 91.5  & 91.4 & 91.3 & 91.2\\
    \end{tabular}
    \caption{Varying k in k-NN}
    \label{tab:topk}
    \end{subtable}
    \hfill
    
	\begin{subtable}[h]{0.5\textwidth}
		\centering
    \centering
    \begin{tabular}{c|cccccc}
         Batch size & 256 & 512 & 1024 & 2048 & 4096 & 8192\\
         \midrule
         Top-1 & 68.7 & 71.7 & 72.9 & 73.5 & \textbf{74.9 }& 74.3 \\
         Top-5 & 88.7 & 90.4 & 91.1 & 91.6 & \textbf{92.1} & 91.9\\
    \end{tabular}
    \caption{Batch size.}
    \label{tab:batch_size}
	\end{subtable}
	\hfill
	\begin{subtable}[h]{0.5\textwidth}
		\centering
		\begin{tabular}{c|ccccc}
         $d$ & 128 & 256 & 512 & 1024 & 2048\\
         \midrule
         Top-1 & 74.9 & 74.9 & 74.8 & 74.9 & 74.6\\
         Top-5 & 92.1 & 92.1 & 92.0 & 92.0 & 92.0\\
    \end{tabular}
    \caption{Varying embedding size $d$}
    \label{tab:embedding_size}
	\end{subtable}
	\hfill
	
	\begin{subtable}[h]{0.5\textwidth}
    \centering
    \begin{tabular}{c|cc}
        Type of NN & Top-1 & Top-5\\
        \midrule
        Soft nearest-neighbor  & 71.4 & 90.4\\
        Hard nearest-neighbor  & {\bf 74.9} & {\bf 92.1}\\

    \end{tabular}
    \caption{Soft vs. hard nearest neighbors as positives.}
    \label{tab:hard_vs_soft}
\end{subtable}
\hfill
\begin{subtable}[h]{0.5\textwidth}
    \centering
    \begin{tabular}{c|cc}
      Prediction MLP & Top-1 & Top-5\\
        \midrule
          &  74.5 & 92.0\\
        \checkmark & {\bf 74.9} & {\bf 92.1}\\        
    \end{tabular}
    \caption{Effect of prediction head.}
    \label{tab:predhead}
\end{subtable}\\

\caption{\textbf{\methodname Ablation Experiments.} Results are obtained for ImageNet linear evaluation.}
\end{table*}

\begin{table}[]
\small
    \centering
    \begin{tabular}{c|c|cc}
        Support set variant & Size & Top-1 & Top-5\\
         \midrule
         NNs from current encoder & 16384 & 74.0 & 91.8 \\
         NNs from queue (older embeddings) & 16384 & 74.2 & 91.7 \\
        
    \end{tabular}
    \caption{\textbf{Different implementations of the support set.}}
    \label{tab:memory_variations}
\end{table}

\noindent \textbf{Compute overhead.} We find increasing the size of the queue results in improved performance but this improvement comes at a cost of additional memory and compute required during training. In Table~\ref{tab:compute} we show how queue scaling with $d=256$ affects memory required during training and number of training steps per second. With a support size of about $98k$ elements we require a modest 100 MB more in memory.

\begin{table}[]
\small
    \centering
    \begin{tabular}{c|ccccc}
         Queue Size & 8192 & 16384 & 32768 & 65536 & 98304\\
         \midrule
         Memory (MB) &  8.4 & 16.8 & 33.6 & 67.3& 100.8\\
         Steps per sec & 6.41 & 6.33 & 6.14 & 5.95 & 5.68  \\
 
    \end{tabular}
    \caption{\textbf{Queue size computational overheads.}}
    \label{tab:compute}
\end{table}

\subsection{Discussion}

\begin{figure}[t]
\begin{center}
   \includegraphics[width=.98\linewidth]{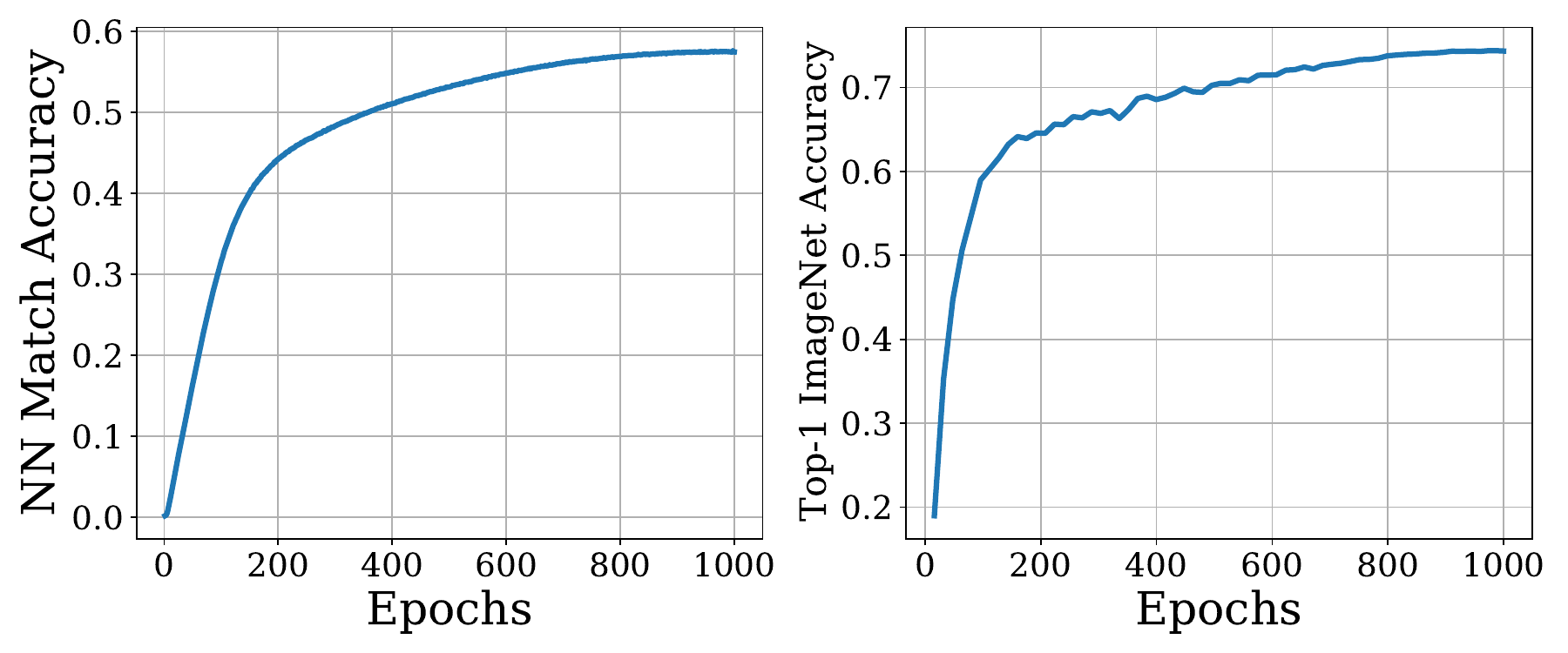}
\end{center}
   \caption{\textbf{NN Match Accuracy vs. Performance.}}
\label{fig:queue_accuracy}
\end{figure}

\noindent \textbf{Ground Truth Nearest Neighbor.} We investigate two aspects of the NN: first, how often does the NN have the same ImageNet label as the query; and second, if the NN is always picked to be from the same ImageNet class (with an Oracle algorithm), then what is the effect on training and the final performance? Figure~\ref{fig:queue_accuracy}, shows how the accuracy of the NN picked from the queue varies as training proceeds. We observe that towards the end of training the accuracy of picking the right neighbor (i.e.\ from the same class) is about $57\%$. The reason that it is not higher is possibly due to random crops being of the background, and thus not containing the object described by the ImageNet label. 

We next investigate if \methodname can achieve better performance if our top-1 NN is always from the same ImageNet class. This is quite close to the supervised learning setup except instead of training to predict classes directly, we train using our self-supervised setup. A similar experiment has also been described as \textit{UberNCE} in \cite{Han20}. This experiment verifies if our training dynamics prevent the model from converging to the performance of a supervised learning baseline even when the true NN is known. To do so, we store the ImageNet labels of each element in the queue and always pick a NN with the same ImageNet label as the query view. We observe that with such a setting we achieve $75.8\%$ accuracy in $300$ epochs. With the Top-1 NN from the support set, we manage to get $72.9\%$ in 300 epochs. This suggests that there is still a possibility of improving performance with a better NN picking strategy, although it might be hard to design one that works in a purely unsupervised way.

\begin{figure}[t]
\begin{center}
   \includegraphics[width=.98\linewidth]{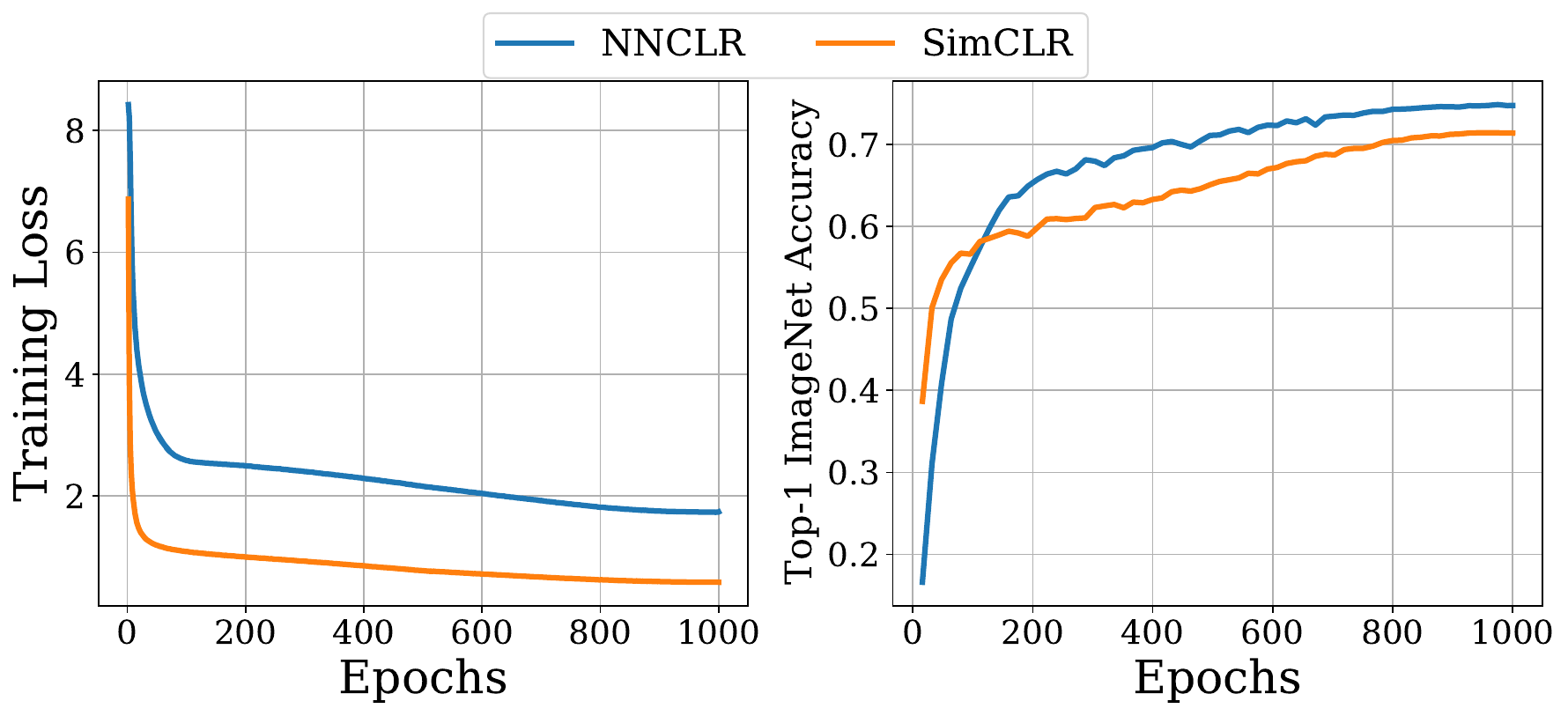}
\end{center}
   \caption{\textbf{\methodname  vs SimCLR} Training curves and linear evaluation curves.}
\label{fig:losscurves}
\end{figure}

\noindent \textbf{Training curves.} In Figure~\ref{fig:losscurves} we show direct comparison between training using cross-entropy loss with an augmentation of the same view as positive (SimCLR) and training with NN as positive (\methodname). The training loss curves indicate \methodname is a more difficult task as the training needs to learn from hard positives from other samples in the dataset. Linear evaluation on ImageNet classification shows that it takes about $120$ epochs for \methodname to start outperforming SimCLR, and remains higher until the end of pre-traininig at $1000$ epochs.

\begin{figure}[t]
\begin{center}
   \includegraphics[width=\linewidth]{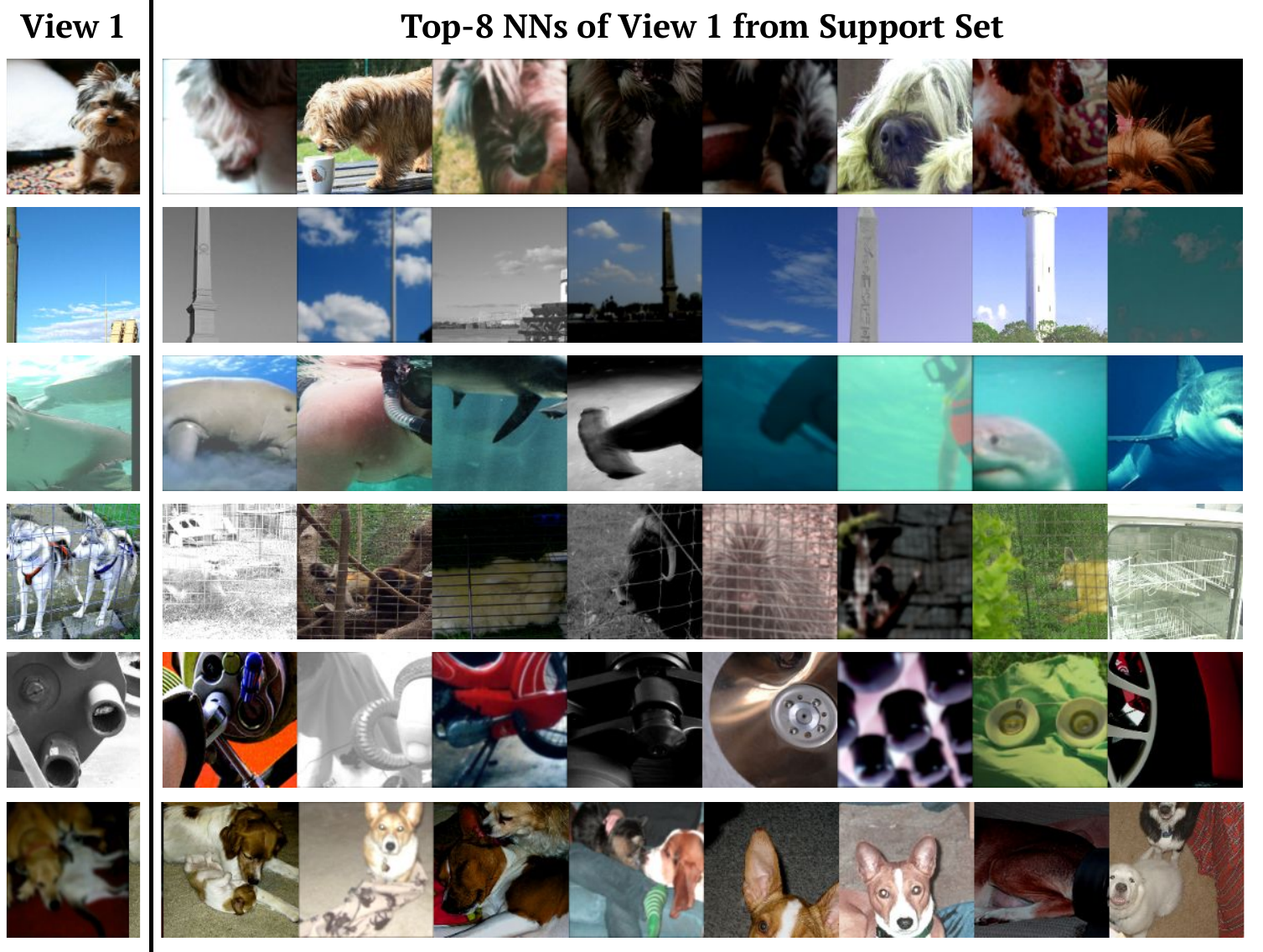}
\end{center}
\caption{\textbf{Nearest neighbors from support set} show the increased diversity of positives in \methodname.}
\label{fig:nn}
\end{figure}

\noindent \textbf{NNs in Support Set.} In Figure~\ref{fig:nn} we show a typical batch of nearest neighbors retrieved from the support set towards the end of training.  Column 1 shows examples of view 1, while the other elements in each row shows the retrieved nearest neighbor from the support set. We hypothesize that the improvement in performance is due to this diversity introduced in the positives, something that is not covered by pre-defined data augmentation schemes. We observe that while many times the retrieved NNs are from the same class, it is not uncommon for the retrieval to be based on other similarities like texture. For example, in row 3 we observe retrieved images are all of underwater images and row 4 retrievals are not from a dog class but are all images with cages in them.

\section{Conclusion}

We present an approach to increase the diversity of positives in contrastive self-supervised learning. We do so by using the nearest-neighbors from a support set as positives. \methodname achieves state of the art performance on multiple datasets. Our method also reduces the reliance on data augmentation techniques drastically.

{\small
\bibliographystyle{ieee_fullname}
\bibliography{egbib}
}

\appendix
\section*{Appendix}

\section{Pseudo-code}
In Algorithm~\ref{alg:code} we present the pseudo-code of \methodname.

\begin{algorithm}[h]
\caption{ Pseudocode}
\label{alg:code}
\definecolor{codeblue}{rgb}{0.25,0.5,0.5}
\definecolor{codekw}{rgb}{0.85, 0.18, 0.50}
\lstset{
  backgroundcolor=\color{white},
  basicstyle=\fontsize{7.5pt}{7.5pt}\ttfamily\selectfont,
  columns=fullflexible,
  breaklines=true,
  captionpos=b,
  commentstyle=\fontsize{7.5pt}{7.5pt}\color{codeblue},
  keywordstyle=\fontsize{7.5pt}{7.5pt}\color{codekw},
}
\begin{lstlisting}[language=python]
# f: backbone encoder + projection MLP
# g: prediction MLP
# Q: queue

for x in loader:  # load a minibatch x with n samples
    x1, x2 = aug(x), aug(x)  # random augmentation
    z1, z2 = f(x1), f(x2)  # projections, n-by-d
    p1, p2 = g(z1), g(z2)  # predictions, n-by-d
    
    NN1 = NN(z1, Q) # top-1 NN lookup, n-by-d
    NN2 = NN(z2, Q) # top-1 NN lookup, n-by-d
    
    loss = L(NN1, p2)/2 + L(NN2, p1)/2 

    loss.backward()  # back-propagate
    update(f, g)  # SGD update
    update_queue(Q, z1) # Update queue with latest projection embeddings

def L(nn, p, temperature=0.1):  
    nn = normalize(nn, dim=1)  # l2-normalize
    p = normalize(p, dim=1)  # l2-normalize
    
    logits = nn @ p.T # Matrix multiplication, n-by-n
    logits /= temperature # Scale by temperature
    
    n = p.shape[0] # mini-batch size
    labels = range(n)
    
    loss = cross_entropy(logits, labels)
    
    return loss
    
def NN(z, Q):
  z = normalize(z, dim=1) # l2-normalize
  Q = normalize(Q, dim=1) # l2-normalize
  sims = z @ Q.T 
  nn_idxes = sims.argmax(dim=1) # Top-1 NN indices
  return Q[nn_idxes]
    
\end{lstlisting}
\end{algorithm}

It is possible to use momentum encoder with NNCLR training. The pseudo-code when momentum encoder is used is shown in Algorithm~\ref{alg:momcode}.
\begin{algorithm}
\caption{Pseudocode with Momentum Encoder}
\label{alg:momcode}
\definecolor{codeblue}{rgb}{0.25,0.5,0.5}
\definecolor{codekw}{rgb}{0.85, 0.18, 0.50}
\lstset{
  backgroundcolor=\color{white},
  basicstyle=\fontsize{7.5pt}{7.5pt}\ttfamily\selectfont,
  columns=fullflexible,
  breaklines=true,
  captionpos=b,
  commentstyle=\fontsize{7.5pt}{7.5pt}\color{codeblue},
  keywordstyle=\fontsize{7.5pt}{7.5pt}\color{codekw},
}
\begin{lstlisting}[language=python]
# f: backbone encoder + projection MLP
# f_m: momentum version of (backbone encoder + projection MLP)
# g: prediction MLP
# Q: queue
# t: tau for momentum encoder

for x in loader:  # load a minibatch x with n samples
    x1, x2 = aug(x), aug(x)  # random augmentation
    z1, z2 = f(x1), f(x2)  # projections, n-by-d
    p1, p2 = g(z1), g(z2)  # predictions, n-by-d
    
    zm1, zm2 = f_m(x1), f_m(x2)  # projections, n-by-d
    
    NN1 = NN(zm1, Q) # top-1 NN lookup, n-by-d
    NN2 = NN(zm2, Q) # top-1 NN lookup, n-by-d
    
    loss = L(NN1, p2)/2 + L(NN2, p1)/2 

    loss.backward()  # back-propagate
    update(f, g)  # SGD update
    update_queue(Q, z_m1) # Update queue with latest projection embeddings from momentum encoder
    
    f_m = t*f_m + (1 - t) * f # Update momentum encoder weights

\end{lstlisting}
\end{algorithm}

\section{Evolution of Nearest-Neighbors}
\begin{figure*}
     \centering
     \begin{subfigure}[b]{0.48\textwidth}
         \centering
         \includegraphics[width=\textwidth]{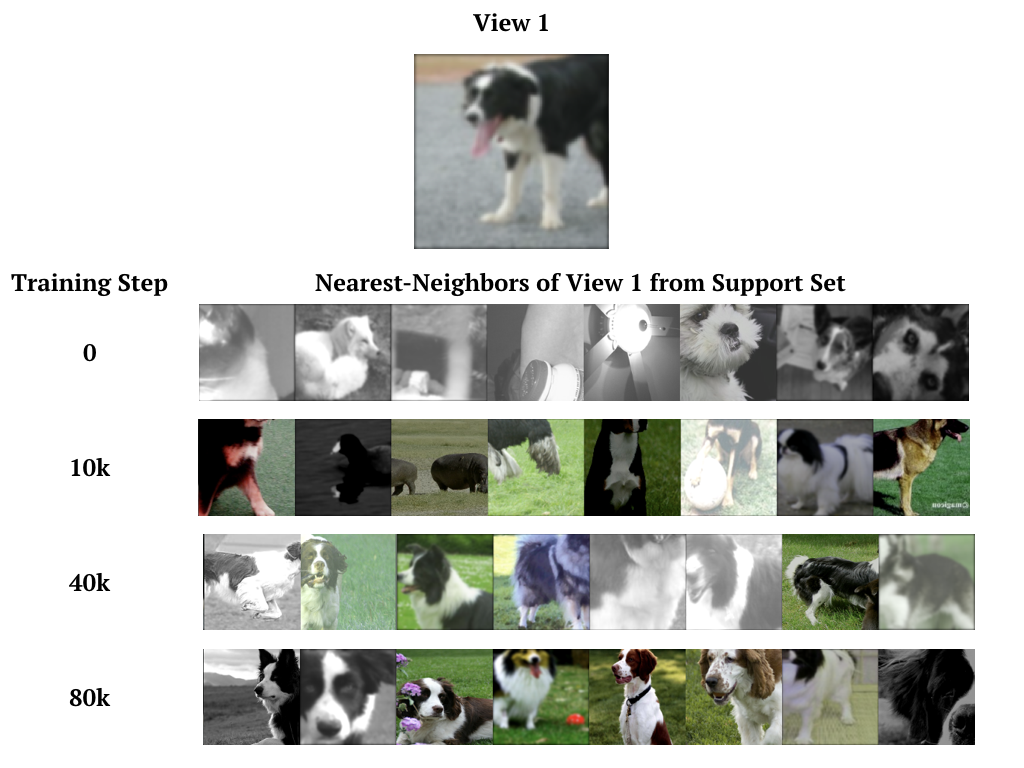}

     \end{subfigure}
     \hfill
     \begin{subfigure}[b]{0.48\textwidth}
         \centering
         \includegraphics[width=\textwidth]{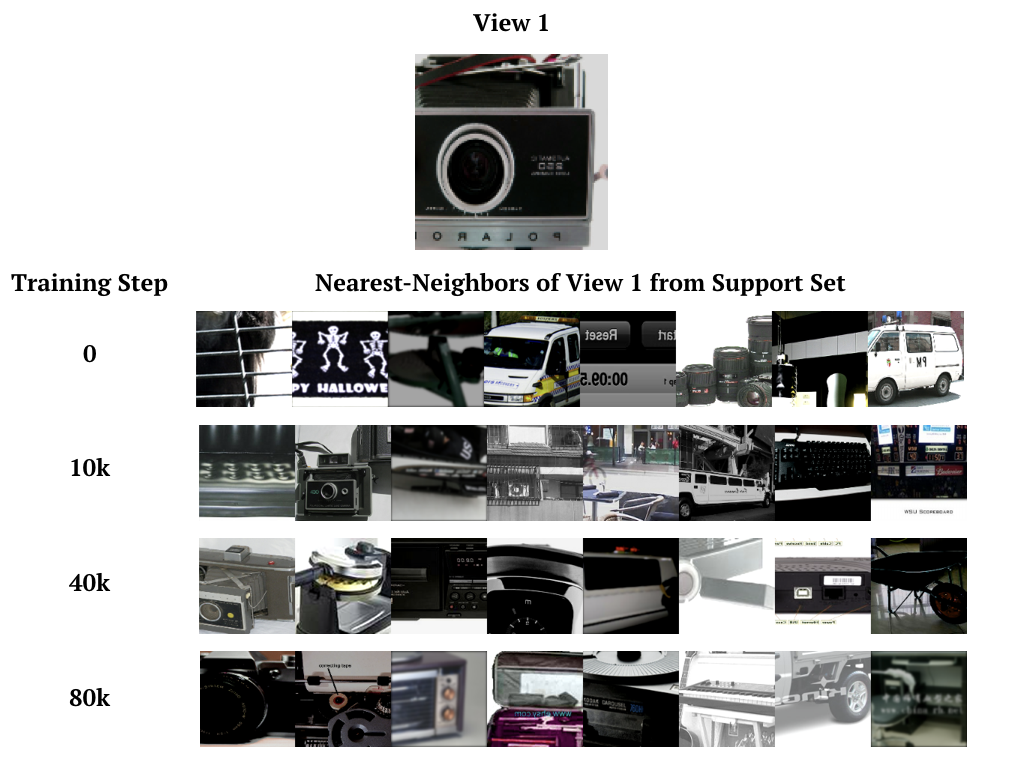}
     \end{subfigure}
     \hfill
     \\
     \begin{subfigure}[b]{0.48\textwidth}
         \centering
         \includegraphics[width=\textwidth]{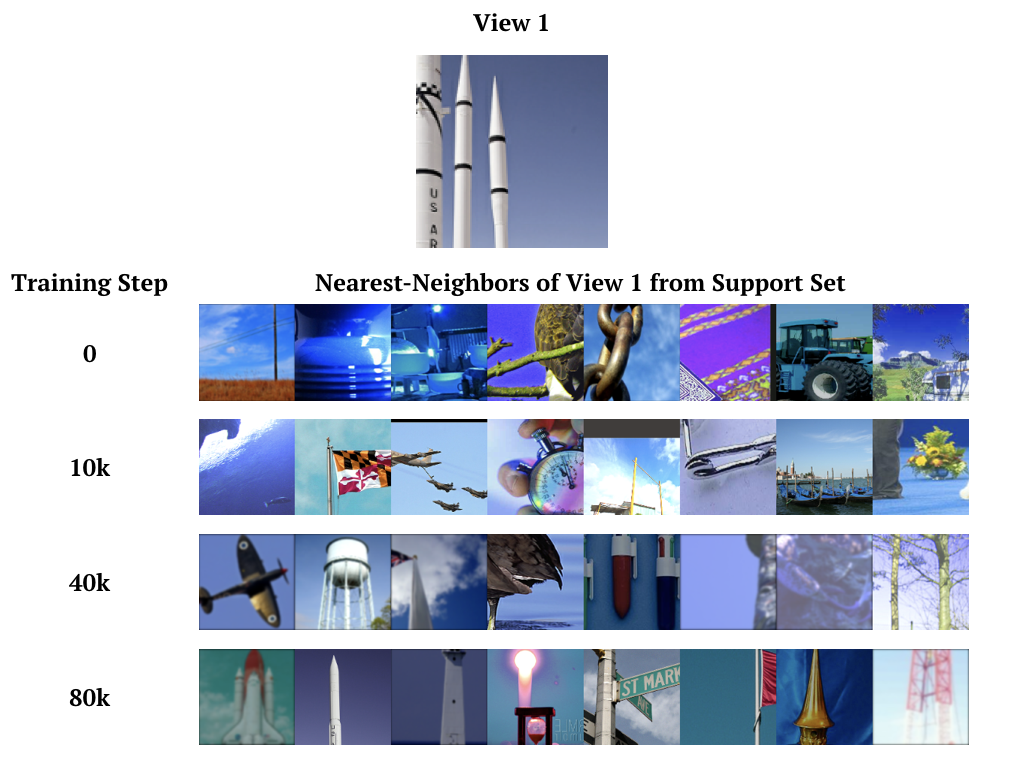}

     \end{subfigure}
     \hfill
     \begin{subfigure}[b]{0.48\textwidth}
         \centering
         \includegraphics[width=\textwidth]{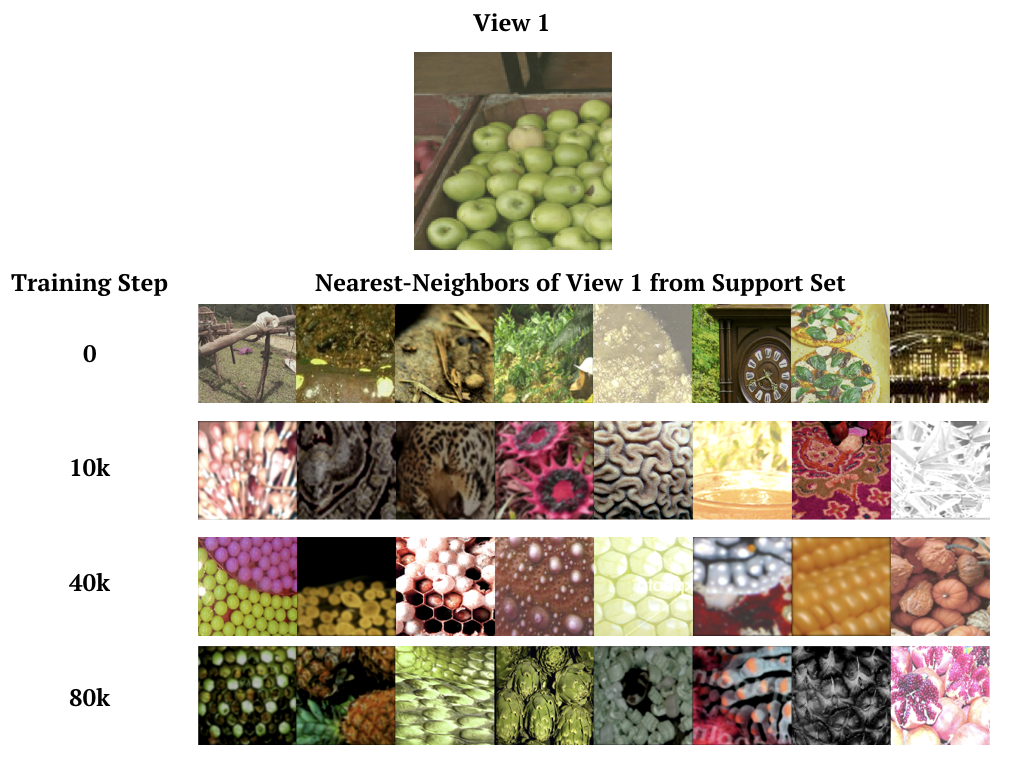}
     \end{subfigure}
        \caption{\textbf{Evolution of Nearest-neighbors} as training proceeds.}
        \label{fig:evolution_nn}
\end{figure*}

In Figure~\ref{fig:evolution_nn} we show how the nearest-neighbors (NN) vary as training proceeds. We observe consistently that in the beginning of training the NNs are usually chosen on the basis of color and texture. As the encoder becomes better at recognizing classes later in training, the NNs tend to belong to similar semantic classes. 

\begin{figure*}[h!]
     \centering
     \begin{subfigure}[b]{0.24\textwidth}
         \centering
         \includegraphics[width=\textwidth]{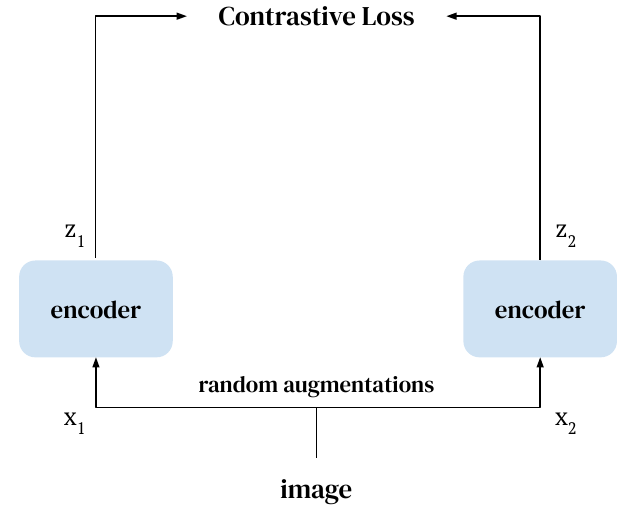}
         \subcaption[]{SimCLR}
     \end{subfigure}
     \hfill
     \begin{subfigure}[b]{0.24\textwidth}
         \centering
         \includegraphics[width=\textwidth]{images/teaser}
         \subcaption[]{\methodname}
     \end{subfigure}
     \hfill
     \begin{subfigure}[b]{0.24\textwidth}
         \centering
         \includegraphics[width=\textwidth]{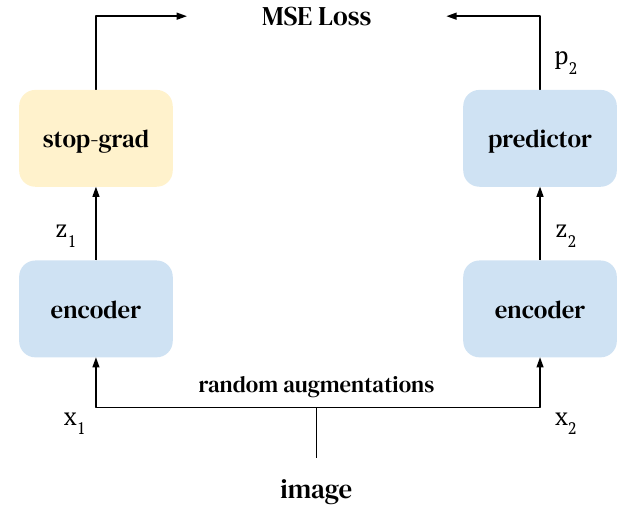}
    \subcaption[]{SimSiam}
     \end{subfigure}
     \hfill
     \begin{subfigure}[b]{0.24\textwidth}
         \centering
         \includegraphics[width=\textwidth]{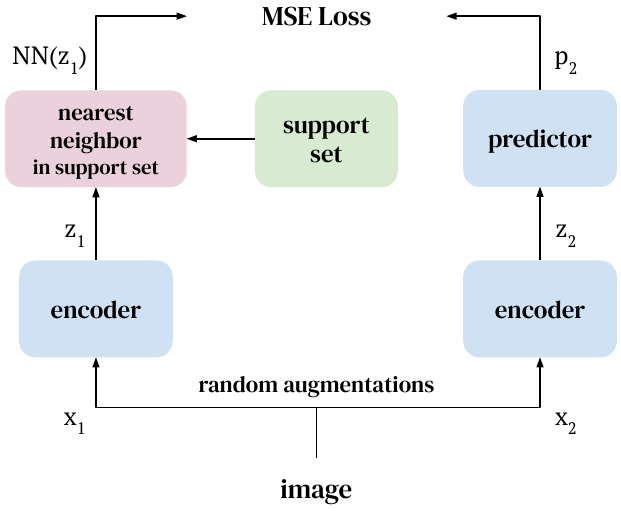}
         \subcaption[]{NNSiam}
     \end{subfigure}
        \caption{\textbf{Comparison of self-supervised methods.}}
        \label{fig:comparison}
\end{figure*}

\section{SimSiam with Nearest-neighbor as Positive}
In this experiment we want to check if it is possible to use the nearest-neighbor in a non-contrastive loss. To do so we use the self-supervised framework SimSiam~\cite{chen2020exploring}, in which the authors use a mean squared error on the embeddings of the 2 views, where one of the branches has a stop-gradient and the other one has a prediction MLP. We replace the stop-gradient branch with its nearest-neighbor from the support set. We call this method NNSiam. In Figure~\ref{fig:comparison} we show how NNSiam differs from SimSiam. We also show how they both differ from SimCLR and \methodname.  Note that there is an implicit stop-gradient in NNSiam because of the use of hard nearest-neighbors. For this experiment, we use an embedding size of 2048, which is the same dimensionality used in SimSiam. We train with a batch size of 4096 with the LARS optimizer with a base learning rate of $0.2$. We find that even with the non-contrastive loss using the nearest neighbor as positive leads to 1.3\% improvement in accuracy under ImageNet linear evaluation protocol. This shows that the idea of using harder positives can lead to performance improvements outside the InfoNCE loss also.

\begin{table}[h]
     \centering
    \begin{tabular}{c|c|cc}
         Method & Positive & Top-1 & Top-5\\
         \midrule
    SimSiam\cite{chen2020exploring} & View 1 & 71.3 & -\\
    NNSiam & NN of View 1 & 72.6 & 90.4 \\
    \end{tabular}
    \caption{\textbf{SimSiam with nearest-neighbor as positive.}}
    \label{tab:simsiam}
\end{table}

\section{Experiments with Vision Transformers}
Vision Transformers (ViT)~\cite{dosovitskiy2020vit} are a class of architectures introduced recently to process images using Transformers. We explore the effectiveness of using self-supervised learning methods to train vision transformers. We find the Adam optimizer to be effective for training ViT models. We train for $1000$ epochs using 2 crops with a base learning rate of $3 \times 10^{-4}$ with a warmup schedule of 10 epochs followed by cosine decay learning schedule, weight decay of $0.05$ and a stochastic depth dropout of $0.1$. We use the output corresponding to the [CLS] token of the final layer as the embedding $z_i$ in the loss and as the representation used as input for the linear classifier. We present the results of this experiment in Table~\ref{tab:vit}. In our experiments, we find NNCLR well suited to train Visual Transformers outperforming the supervised learning model (trained without the augmentation introduced in DeIT~\cite{touvron2021training}) by $4.4\%$ and the SimCLR model by $2\%$.

\begin{table}[h]
\small
     \centering
    \begin{tabular}{c|c|c}
         Arch. & Training Loss &  Top-1 \\ 
         \midrule
         ViT-B/16 & Supervised Learning & 72.1\\
         ViT-B/16~\cite{touvron2021training} & Supervised Learning & \textbf{81.8}\\
         \midrule
         ViT-B/16 & SimCLR & 74.5\\
         ViT-B/16 & NNCLR & \textbf{76.5}\\
    \end{tabular}
    \caption{\textbf{Vision Transformer} Top-1 ImageNet accuracy under the linear protocol. }
    \label{tab:vit}
\end{table}

\section{Self-supervised Learning as a Pre-training Step for Supervised Learning}
\label{sec:sslpt}
Until recently it was believed supervised learning on a particular dataset would always be better than self-supervised learning on that dataset. BYOL~\cite{grill2020bootstrap} showed that some large models can end up achieving higher accuracy than their supervised counterparts. In this experiment, we similarly show that self-supervised learning can serve as a useful pre-training step for supervised learning, alleviating the need for extra data or complex regularization techniques that are used to boost the performance of the final model.

In this experiment we first pretrain a model with NNCLR for 1000 epochs on the ImageNet 2012 dataset. We then proceed to perform regular supervised training for 100 epochs. This is similar to the semi-supervised learning setup but with 100\% of the labels available for fine-tuning the pre-trained model. In Table~\ref{tab:ssl_pt_sl} we show results on this experiment. We observe initialization from a self-supervised model improves the performance of ResNet50 from 76.2\% to 79.1\%. This performance is comparable to training a ResNet-50 model with JFT-300M dataset~\cite{kolesnikov2020big} which is considerably more data than ImageNet. We find this pre-training technique is especially useful with the newly proposed ViT~\cite{dosovitskiy2020vit} architecture which requires additional data or regularization techniques to achieve good performance. ViT-B/16 pre-trained with NNCLR outperforms DeIT~\cite{touvron2021training} but is only worse than the same architecture trained with the JFT-300M dataset by 0.5\%. This experiment highlights another use-case for self-supervised learning: to provide a good initialization for supervised learning.

\begin{table}[h]
\small
     \centering
    \begin{tabular}{cccc|c}
         Arch. & Res. & SS PT & Reg.
         / Extra Data & Top-1 \\ 
         \midrule
         ResNet50~\cite{he2016deep} & 224 & - & - & 76.2 \\
         ResNet50~\cite{kolesnikov2020big} & 224& - & JFT-300M & 79.0 \\
         ResNet50 & 224& NNCLR & - & \textbf{79.1} \\
         \midrule
         
    ViT-S/16 & 224 & - &  DeIT Aug.& 79.8 \\
         ViT-S/16 & 224 & NNCLR &  DeIT Aug.& \textbf{82.7}\\
         \midrule
         ViT-B/16~\cite{touvron2021training} & 224 & - & DeIT Aug. & 81.8\\
         ViT-B/16 & 224 & NNCLR & DeIT Aug.& \textbf{82.5}\\
         \midrule
         
         ViT-B/16~\cite{dosovitskiy2020vit} & 384 & - & - & 77.9 \\
         ViT-B/16~\cite{touvron2021training} & 384 & - & DeIT Aug. & 83.1\\
         ViT-B/16~\cite{dosovitskiy2020vit} & 384 & - & JFT-300M & \textbf{84.2} \\
         ViT-B/16 & 384 & NNCLR & DeIT Aug. & 83.7 \\
         \midrule
          ViT-B/8 & 224 & - & DeIT Aug. & 83.1 \\
         ViT-B/8 & 224 & NNCLR &  DeIT Aug. & \textbf{84.4}\\
    
    \end{tabular}
    \caption{\textbf{Self-supervised learning as pre-training for supervised learning.} Top-1 ImageNet accuracy. Both self-supervised and supervised training are done on ImageNet 2012 training set. Arch: Architecture, SS PT: Self-supervised Pre-training Technique, Reg. Additional regularization techniques}
    \label{tab:ssl_pt_sl}
\end{table}

\section{Transfer Learning}
In this section we study the performance of Visual Transformers (ViT) in the transfer learning setting. To do this, we repeat the experiment described in Section~\ref{sec:transfer_learning} with ViT models. The results of this experiment are presented in Table~\ref{tab:transfer_learning_vit}. First, we observe that ViT-B/16 trained with only a supervised learning objective using the data augmentation strategy outlined in DeIT\cite{touvron2021training} on the ImageNet dataset transfers poorly as compared to ResNet50 architecture trained with the supervised loss. ViT-B/16 is worse on 10 datasets out of the 12 datasets in the transfer learning benchmark. However, ViT-B/16 trained with \methodname objective outperforms the ResNet50 architecture trained with the same loss on 8 out of the 12 datasets in the benchmark. Additionally, ViT-B/16 trained with \methodname has better performance on all datasets as compared to the same model trained with just the supervised loss. This shows that the Vision Transformer's potential for transfer learning is enhanced by using self-supervised training like \methodname. Of particular note is the increase in performance on Birdsnap ($\sim10.7\%$), Sun397 ($\sim5.8\%$), Cars ($\sim8.0\%$), Aircraft ($\sim 13.2\%$), DTD ($\sim5.8\%$) and Flowers ($\sim9.3\%$). We also train a ViT-B/8 model that has the same architecture as ViT-B/16 but uses $8\times8$-sized non-overlapping patches in the images to produce the tokens used in the Transformer architecture. We find that ViT-B/8 outperforms ViT-B/16 on all datasets in the transfer learning setup. We also observe that \methodname brings significant gains over just supervised learning. Of particular note is the increase in performance on Birdsnap ($\sim9.8\%$), Sun397 ($\sim3.9\%$), Aircraft ($\sim 10.5\%$), DTD ($\sim7.1\%$) and Flowers ($\sim5.7\%$). Finally, we test the transfer learning performance on models pre-trained with \methodname and fine-tuned with the supervised loss as outlined in Section~\ref{sec:sslpt}. We find this setup increases the performance over the already strong baseline of ViT-B/8 trained with just \methodname. We find a boost of 8.3\% on Food, 5.3\% on Birdsnap, 2.9\% on Sun397, 15.8\% on Cars, 2.2\% on VOC2007, 2.1\% on Pets. However, we also observe fine-tuning on the supervised loss also results in degradation in performance on some datasets: 1.1\% on DTD and 1.1\% on Flowers. Inspite of the degradation in performance, ViT-B/8 pretrained with \methodname and fine-tuned with the supervised loss on ImageNet is always better than just using the supervised loss only. Overall, we find Vision Transformers are well suited for transfer learning if they have been pre-trained with a self-supervised loss like \methodname.

\begin{table*}[]
\footnotesize
    \centering
    \begin{tabular}{l|c|c|c|c|c|c|c|c|c|c|c|c|c}
         Method & Arch. & Food & CIFAR10 & CIFAR100 &  Birdsnap & SUN397 &  Cars & Aircraft & VOC2007 &  DTD & Pets & Cal.101 & Flowers\\
         \midrule
         BYOL & R50 &  75.3 & 91.3 &  78.4  & 57.2 & 62.2   & 67.8 &  60.6& 82.5 & 75.5 & 90.4 & 94.2 & 96.1 \\
         SimCLR & R50 & 72.8 & 90.5 & 74.4 & 42.4 & 60.6 & 49.3  & 49.8 & 81.4 & 75.7 & 84.6 & 89.3 & 92.6 \\
         Sup.-IN & R50 & 72.3 & 93.6 & 78.3 &53.7 & 61.9 & 66.7 & 61.0& 82.8 & 74.9 & 91.5 & \textbf{94.5} & 94.7 \\
         \methodname & R50& 76.7  & 93.7 & 79.0  & 61.4 & 62.5  & 67.1 & 64.1 & 83.0 & 75.5  & 91.8 & 91.3 & 95.1\\
         \midrule
          Sup. IN & ViT-B/16 & 71.4 & 95.2 & 80.5 & 52.8 & 59.3 & 50.7 & 52.1 & 81.2 & 71.5 & 91.4 & 85.6 & 85.6 \\
          \methodname & ViT-B/16 & 72.7 & 95.7 & 82.3 & 63.5 & 65.5 & 58.3 & 65.3 & 83.2 & 77.3 & 92.5 & 88.5 & 94.9 \\
          \midrule
          Sup. IN & ViT-B/8 & 76.2 & 95.9 & 81.4 & 61.9 & 62.6 & 61.4 & 58.6 & 83.8 & 72.1 & 93.2 & 85.5 & 90.1\\
         \methodname & ViT-B/8 & 77.2 & \textbf{96.8} & 82.0 & 71.7 & 66.5 & 58.6 &\textbf{ 69.1 }& 84.2 & \textbf{79.2} & 92.8 & 88.7 & \textbf{95.8}\\
          \methodname + Sup & ViT-B/8 & \textbf{85.5}& 96.2 & \textbf{84.7 }& \textbf{77.0} & \textbf{69.4} & \textbf{74.4} & 68.8 & \textbf{86.4} & 78.1 &\textbf{ 94.9} & 90.5 & 94.7\\
    \end{tabular}
    \caption{\textbf{Transfer learning performance} using ResNet-50 and ViT-B/16 pretrained with ImageNet. For all datasets we report Top-1 classification accuracy except Aircraft, Caltech-101, Pets and Flowers for which we report mean per-class accuracy and VOC2007 for which we report 11-point MAP.}
    \label{tab:transfer_learning_vit}
\end{table*}

\section{Visualizations}

\subsection{Attention in NNCLR ResNet}
\label{sec:resnet_attention}
\begin{figure*}
     \centering
     \begin{subfigure}[b]{0.49\textwidth}
         \centering
         \includegraphics[width=\textwidth]{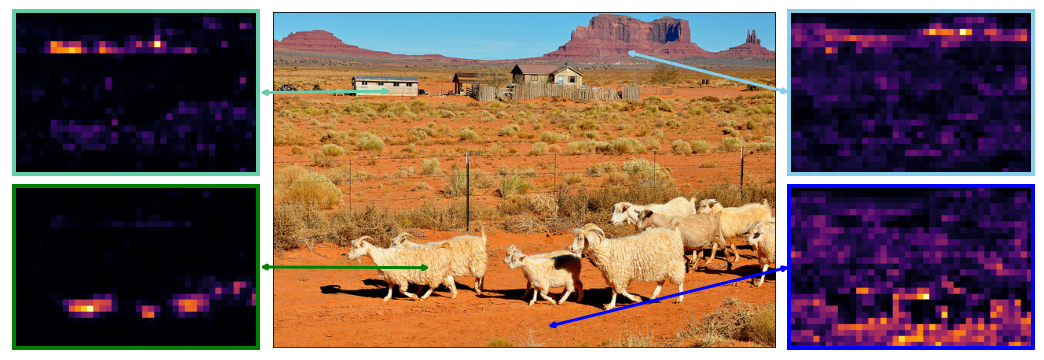}
         \caption{House, sheep, mountains, ground}
     \end{subfigure}
     \hfill
     \begin{subfigure}[b]{0.49\textwidth}
         \centering
         \includegraphics[width=\textwidth]{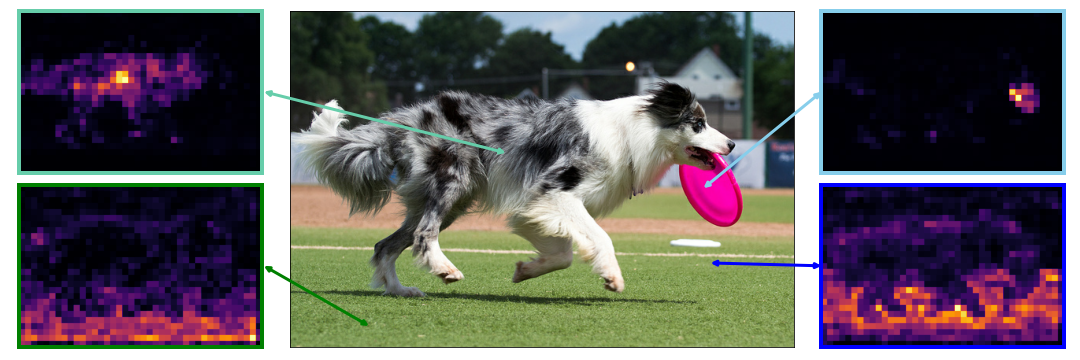}
         \caption{Dog, grass, frisbee, grass}
     \end{subfigure}
     \hfill
     \begin{subfigure}[b]{0.49\textwidth}
         \centering
         \includegraphics[width=\textwidth, height=1.32in]{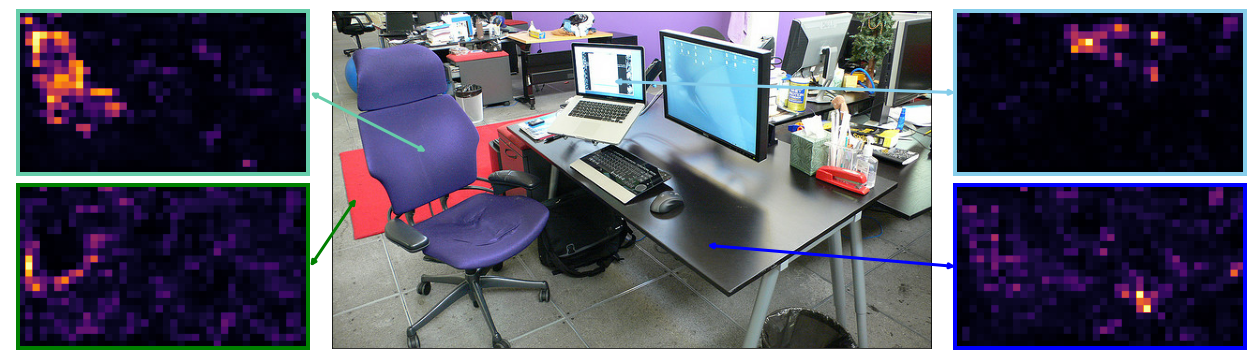}
         \caption{Chair, mat, monitor, table}
     \end{subfigure}
     \hfill
     \begin{subfigure}[b]{0.49\textwidth}
         \centering
         \includegraphics[width=\textwidth]{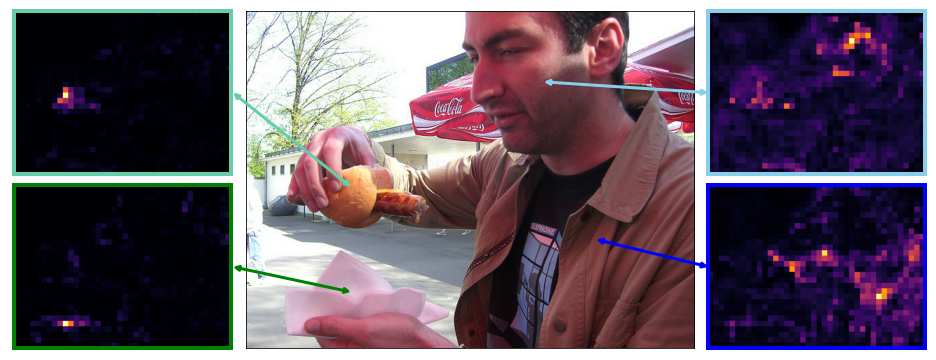}
         \caption{Hotdog, napkin, face, jacket}
     \end{subfigure}
     \hfill
     \begin{subfigure}[b]{0.49\textwidth}
         \centering
         \includegraphics[width=\textwidth]{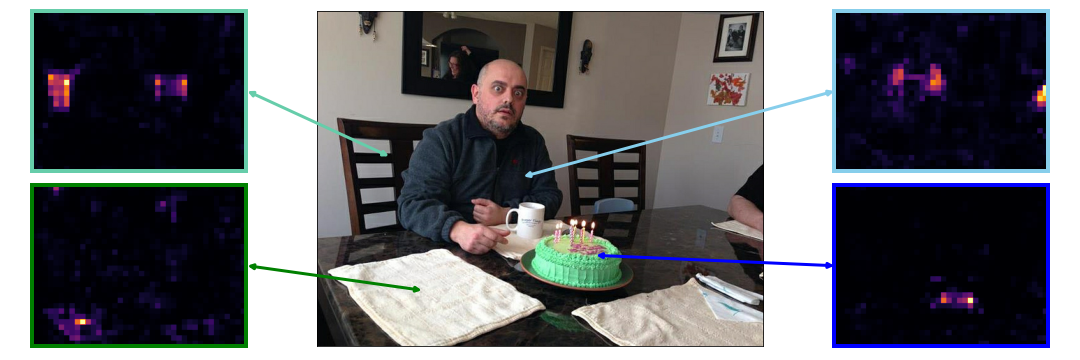}
         \caption{Chair, napkin, person, cake}
     \end{subfigure}
     \hfill
     \begin{subfigure}[b]{0.49\textwidth}
         \centering
         \includegraphics[width=\textwidth]{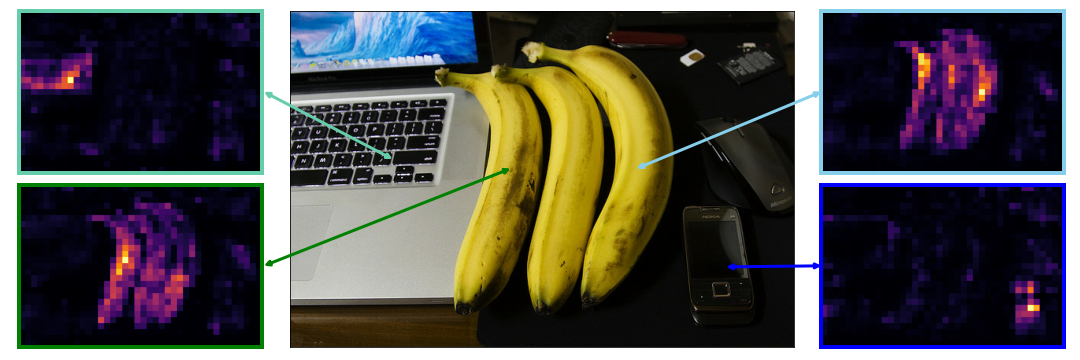}
         \caption{Keyboard, banana, banana, mobile phone}
     \end{subfigure}
     \hfill
     
          \caption{\textbf{Visualization of Attention in NNCLR ResNets.} We observe that self-supervised pre-training with NNCLR results in localization of objects of different categories.}
          \label{fig:resnet_point_attention}

\end{figure*}

We visualize ``attention" of various models trained with NNCLR to probe what the model might be focusing on. In order to visualize attention, we need a query embedding and the feature map produced by the image. We convolve the query embedding across the feature map to produce an attention map. In Figure~\ref{fig:resnet_point_attention}, we show examples of attention corresponding to different query embeddings on images from the COCO dataset. We use the original resolution of the images to get a larger feature map that can highlight object boundaries and locations of small objects. Each arrow points to the query embedding in the image and the corresponding attention of that query embedding in the feature map. In each sub-figure's caption we also mention the object class they are pointing at. In \cite{caron2021emerging}, the authors observe that vision transformer models trained with self-supervised losses like DINO~\cite{caron2021emerging} show emergence of properties like objectness and part localization in the attention layer of the transformer. We find these emerging properties also exist in ResNet50 models trained with self-supervised losses like NNCLR. 

\subsection{Cross-image Attention with NNCLR ResNets}
\label{sec:cross_attention}
\begin{figure*}
     \centering
     \begin{subfigure}[ht]{\textwidth}
         \centering
         \includegraphics[width=\textwidth]{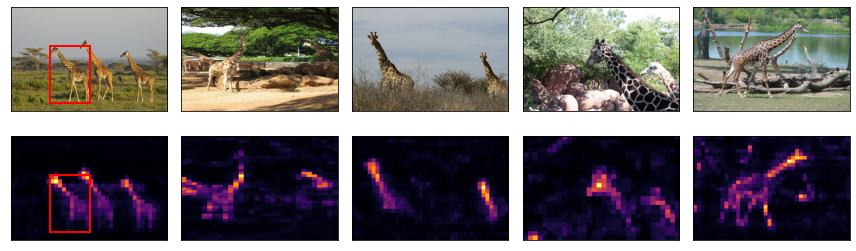}
         \caption{Giraffe}
     \end{subfigure}
     \begin{subfigure}[ht]{\textwidth}
         \centering
         \includegraphics[width=\textwidth]{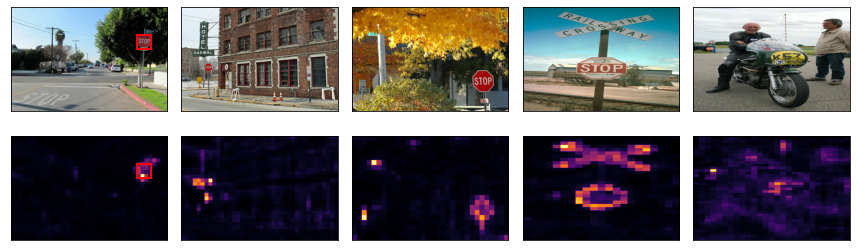}
         \caption{Stop-sign}
     \end{subfigure}
     \begin{subfigure}[ht]{\textwidth}
         \centering
         \includegraphics[width=\textwidth]{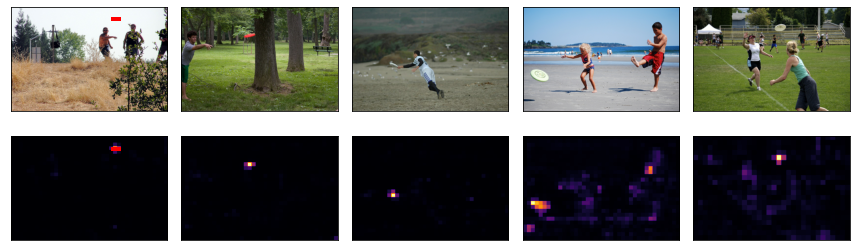}
         \caption{Frisbee}
     \end{subfigure}
     \begin{subfigure}[ht]{\textwidth}
         \centering
         \includegraphics[width=\textwidth]{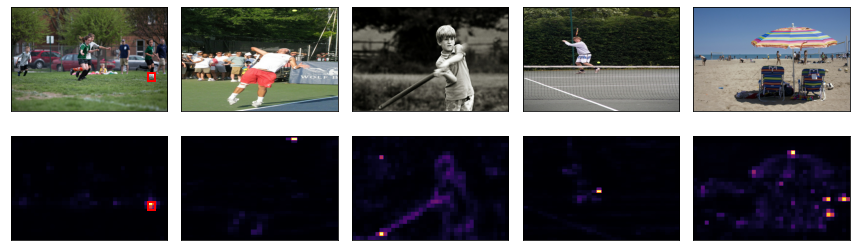}
         \caption{Ball}
     \end{subfigure}
     \caption{\textbf{Cross-Attention.} We use features of an object in the images in the first column (red box in leftmost column shows the query object) to look for the same object in images in other columns. We find NNCLR features of the same object in different images are close to each other.}
     \label{fig:cross_attention}
\end{figure*}

In the previous section we presented attention of a query embedding with  parts of the same image. To visualize if models trained with NNCLR encode object categories, we conduct the following experiment. We use a query embedding of an object from one image by average pooling the features in the bounding box of an object denoted by the red box. We convolve this query embedding over the feature maps obtained by passing other images through a ResNet50 trained with \methodname loss. We show the results of this experiment in Figure~\ref{fig:cross_attention}. We observe that features of the same object in different images are close to each other in the NNCLR feature space. The results of this experiment highlight that the learned features encode semantic similarity beyond color, in order to localize objects of the same category across different images as shown in Figure~\ref{fig:cross_attention}. In the first row, the features are able to localize multiple objects  of the same category \textit{giraffe}. In the second row, we show that while the ground-truth class is that of \textit{stop-sign} the features are considering different kinds of street signs (hotel name, street name sign, railroad-crossing sign) as similar. This is interesting because the query features are pooled only from the stop-sign region but the retrieved regions of similarity are of different colors. We also observe the model does not activate over the text of the stop-sign. In the last two rows, we show how the model is able to localize small objects (like ball and frisbee) in different images.

\subsection{Comparison of Attention in ResNets vs ViT}
\label{sec:comparison_attention}
\begin{figure*}
     \centering
    \includegraphics[width=\textwidth]{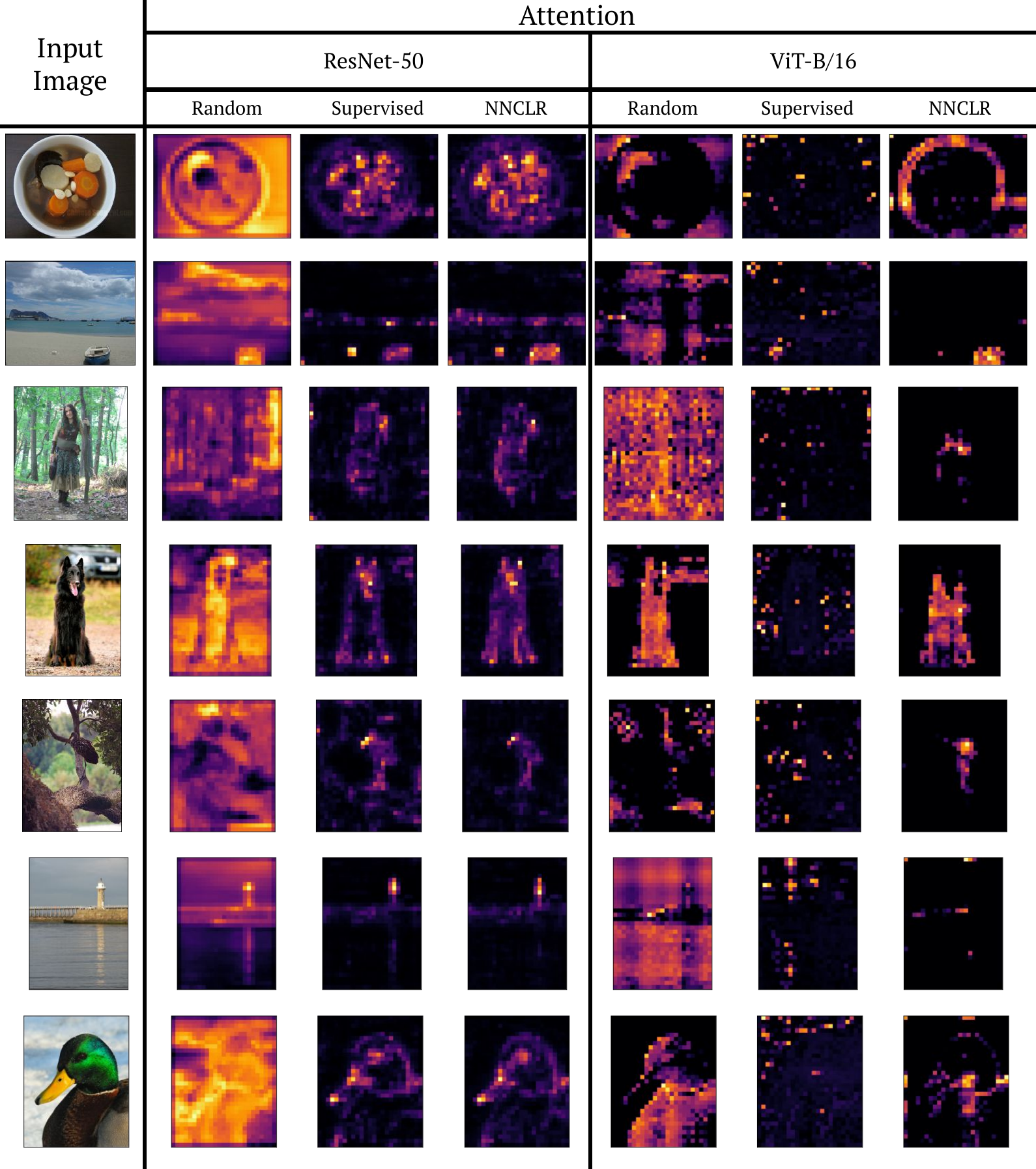}
     
     \caption{\textbf{Comparison of Attention Maps.} We compare attention between feature map and the global image embedding of the last layer of ResNet-50 and ViT-B/16 architectures with three different sets of weights, \textit{Random}: randomly initialized weights, \textit{Supervised}: weights after training a model with supervised learning loss, \textit{Supervised}: weights after training a model with the \methodname objective. More details in Section~\ref{sec:comparison_attention}}. 
     \label{fig:rvsv}
\end{figure*}

In this experiment we want to compare the attention maps of 2 architectures, ResNet-50 and ViT-B/16, each with 3 different weights: randomly initialized, ImageNet supervised and ImageNet NNCLR. For ViT we use the average self-attention over all heads in the final layer for the [CLS] token. For ResNet-50 we use the method described in Sec.~\ref{sec:resnet_attention} and use the average pooled embedding as the query embedding since it is the equivalent of the [CLS] token in ViTs. We show our results in Figure~\ref{fig:rvsv}. First we observe the attention maps delineate salient objects in the image. Similar to DINO~\cite{caron2021emerging} we observe that ViTs trained with just a supervised learning objective do not have objects highlighted in their attention map. However, we do observe that both supervised and self-supervised ResNets have delineated objects in their attention maps. We also note that sometimes randomly initialized ResNets (rows 2 and 5) and ViTs (rows 4 and 7) are able to localize individual objects in their attention map because of similarity in color and texture across the spatial extent of the object. This fact makes it difficult to conclude that a model that produces the well-delineated objects in the self-attention map will necessarily have learned semantically meaningful features. We suggest using a combination of self-attention and  cross-attention maps (described in Section~\ref{sec:cross_attention}) as a more robust visualization technique for interpreting semantic features.

\end{document}